%% file: main_arxiv.tex
\documentclass[10pt,twocolumn,letterpaper]{article}

\usepackage{iccv}
\usepackage{times}
\usepackage{epsfig}
\usepackage{graphicx}
\usepackage{amsmath}
\usepackage{amssymb}

\usepackage{xspace} 
\include{def} 
\usepackage{subcaption} 
\usepackage{booktabs}
\usepackage{xcolor}
\usepackage{multirow} 
\usepackage{bigstrut} 
\usepackage{algorithm} 
\usepackage{algorithmic} 
\usepackage{dsfont} 
\usepackage{pifont} 
\usepackage{colortbl} 
\usepackage{arydshln}
\usepackage{pdfpages} 

\usepackage[pagebackref=true,breaklinks=true,letterpaper=true,colorlinks,bookmarks=false]{hyperref}

\iccvfinalcopy 

\def\iccvPaperID{1987} 

\ificcvfinal\fi

\def\shortname{CPCM\xspace}
\def\name{Contextual Point Cloud Modeling\xspace}
\def\lowername{contextual point cloud modeling\xspace}

\newcommand\rotateboxtable[2][0]{\rotatebox{#1}{#2}}
\newcommand\invisiblesection[1]{\noindent\textbf{#1}.}
\newcommand{\revision}{\textcolor{black}}

\begin{document}

\title{\shortname: \name for Weakly-supervised \\ Point Cloud Semantic Segmentation}

\author{Lizhao Liu$^{1,2}$~~Zhuangwei Zhuang$^{1,2}$~~Shangxin Huang$^1$~~Xunlong Xiao$^1$~~Tianhang Xiang$^{1}$ \\
Cen Chen$^{1}$~~Jingdong Wang$^{3}$~~~Mingkui Tan$^{1,2}$\footnotemark[2]\\
$^1$South China University of Technology~~$^2$Pazhou Lab~~$^3$Baidu Inc.\\
{\tt\small \{selizhaoliu, z.zhuangwei, sevtars, sexxl, sexiangtianhang\}@mail.scut.edu.cn,} \\ {\tt\small \{chencen, mingkuitan\}@scut.edu.cn, wangjingdong@baidu.com}
}


\maketitle
\ificcvfinal\thispagestyle{empty}\fi

\renewcommand{\thefootnote}{\fnsymbol{footnote}}
\footnotetext[2]{Corresponding author.}
\renewcommand{\thefootnote}{\arabic{footnote}}

\begin{abstract}
   We study the task of weakly-supervised point cloud semantic segmentation with sparse annotations (\eg less than 0.1\% points are labeled), aiming to reduce the expensive cost of dense annotations. 
   Unfortunately, with extremely sparse annotated points, it is very difficult to extract both contextual and object information for scene understanding such as semantic segmentation. 
   Motivated by masked modeling  (\eg~MAE) in image and video representation learning,
   we seek to endow the power of masked modeling to learn contextual information from sparsely-annotated points. However, directly applying MAE to 3D point clouds with sparse annotations may fail to work. 
   First, it is non-trivial to effectively mask out the informative visual context from 3D point clouds.
   Second, how to fully exploit the sparse annotations for context modeling remains an open question. 
   In this paper, we propose a simple yet effective \name (\shortname) method that consists of two parts: a region-wise masking (RegionMask) strategy and a contextual masked training (CMT) method. Specifically, RegionMask masks the point cloud continuously in geometric space to construct a meaningful masked prediction task for subsequent context learning. CMT disentangles the learning of supervised segmentation and unsupervised masked context prediction for effectively learning the very limited labeled points and mass unlabeled points, respectively.
   Extensive experiments on the widely-tested ScanNet V2 and S3DIS benchmarks demonstrate the superiority of \shortname over the state-of-the-art.
\end{abstract}


\begin{figure}[!ht]

    \begin{subfigure}{0.48\textwidth}
    \centering
    \includegraphics[width=0.94\linewidth]{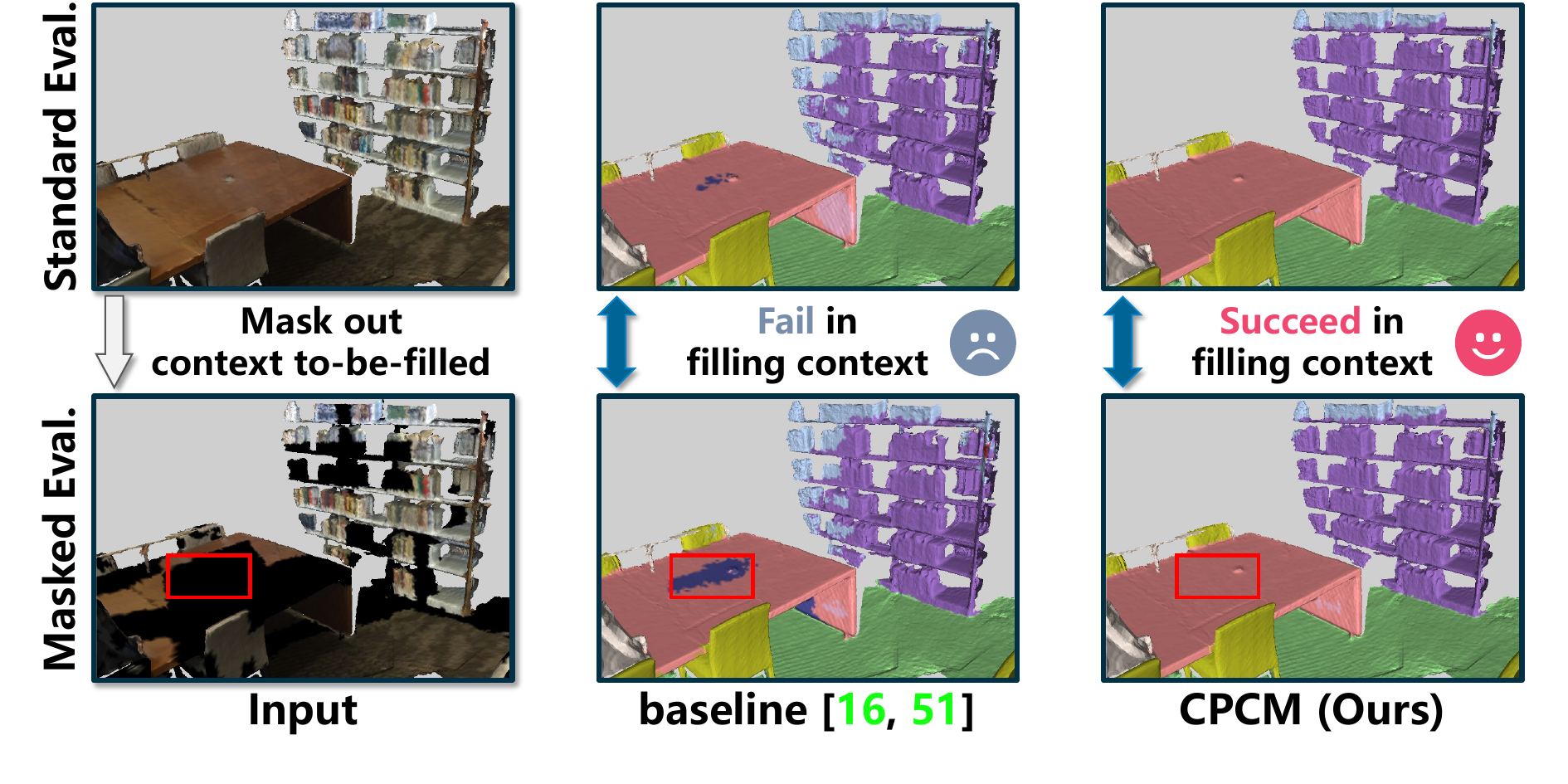}
    \end{subfigure}

    \begin{subfigure}{0.48\textwidth}
    \centering
    \includegraphics[width=0.94\linewidth]{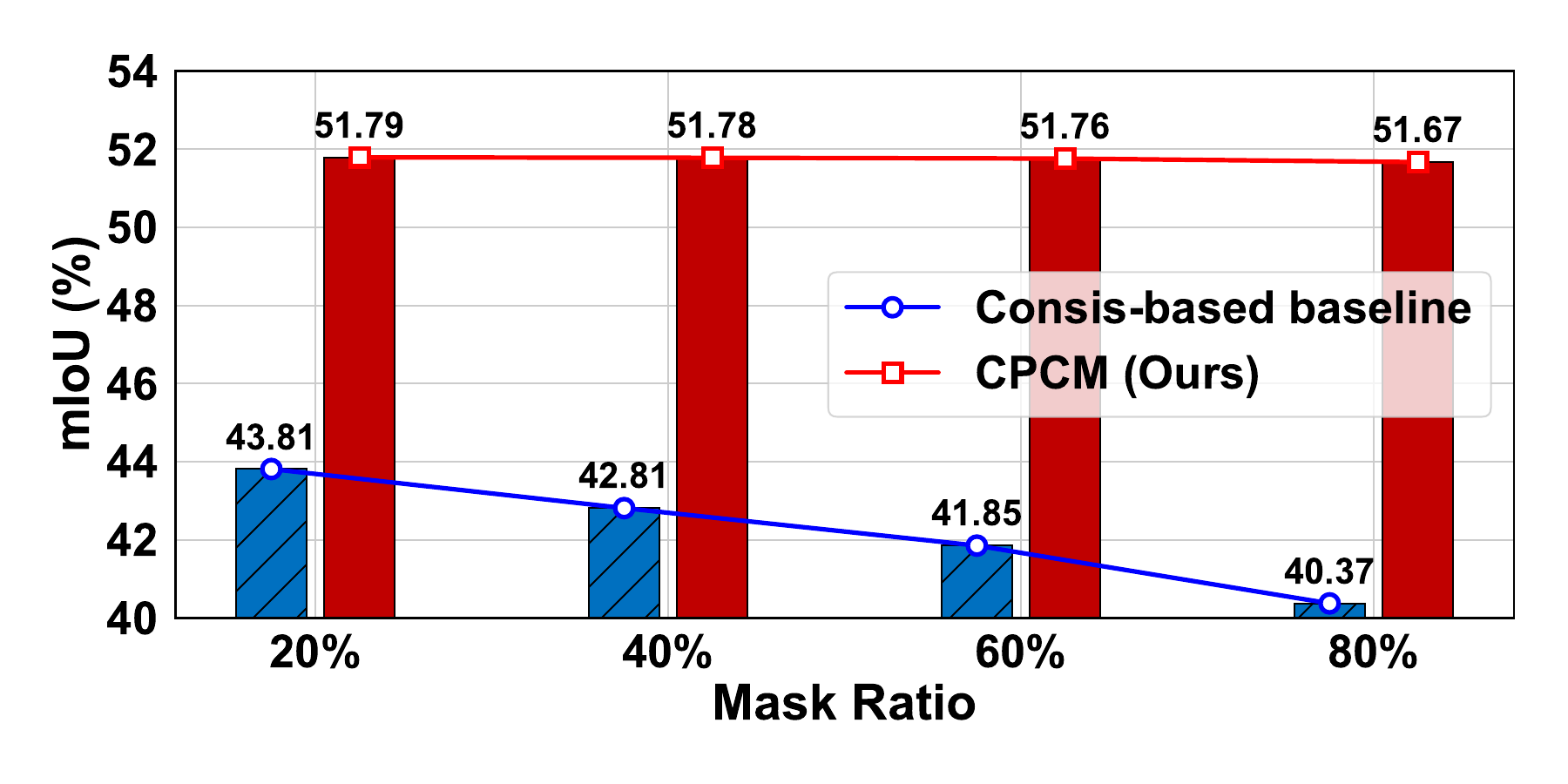}
    \end{subfigure}
    
   \caption{Effectiveness of the proposed \shortname on context comprehension ability compared to the consistency-based baseline~\cite{li2022hybridcr,zhang2021perturbed}. We conduct masked evaluations to inspect the model's contextual understanding ability.
    \revision{The visual comparison of results from different methods (mask ratio = 40\%) and the performance \wrt different mask ratios are shown in the top and bottom panels, respectively.}
    }
    \vspace{-20 pt}
    \label{fig:motivation}

\end{figure}

\section{Introduction}
\label{sec:intro}
With the growing demand for autonomous driving and robotic navigation, point cloud semantic segmentation becomes an indispensable technique for accurate 3D environment perception~\cite{li2020deep,nguyen20133d,zermas2017fast}. Recent years have witnessed great progress in fully-supervised learning in point cloud segmentation~\cite{cheng20212,graham20183d,hu2020randla,le2018pointgrid,qi2017pointnet,qi2017pointnet++,tatarchenko2018tangent,xu2021rpvnet,zhuang2021perception}. {However, densely-annotating point-wise labels are time-consuming, labor-intensive as well as economic-inefficient to obtain since the number of points in point cloud data can easily reach tens of thousands of magnitude~\cite{wei2020multi,xu2020weakly}.} It goes without saying that diving into point cloud semantic segmentation from sparse labels is crucial to
reduce the annotation cost and expand the application boundary~\cite{hu2022sqn,liu2020dynamic,liu2022less}.


Very recently, to reduce the reliance on dense labels while still delivering satisfactory point cloud semantic segmentation performance, most effort has been put into learning from the weakly-annotated labels~\cite{hu2022sqn,li2022hybridcr,liu2021one,wei2020multi,xu2020weakly,yang2022mil,zhang2021weakly,zhang2021perturbed}. {Among several types of weakly-annotated labels, the partial point-wise labeling scheme offers the best trade-off between annotation cost and segmentation performance~\cite{hu2022sqn,liu2022less}.}
In the partially annotated point cloud data, the labeled part typically occupies a very small portion of points (\eg 0.1\%) per scene~\cite{hu2022sqn}. In this case, directly applying supervised cross-entropy loss {only} on the limited labeled part is prone to overfitting~\cite{liu2021one,ren20213d,wei2021dense}. {As a result, the primary challenge is learning from a significant proportion of unlabeled points to improve model generalization performance, rather than utilizing only the labeled points~\cite{li2022hybridcr,zhang2021perturbed}.} 


Existing methods seek to tackle the challenge by exploiting different levels of feature consistency under various data augmentations.
To be specific, researchers resort to enforcing feature consistency between differently augmented or geometrically calibrated point clouds by discriminating points from different scenes with contrastive learning~\cite{hou2021exploring,jiang2021guided,li2022hybridcr,xie2020pointcontrast}, exploring color \& geometric smoothness~\cite{xu2020weakly,zhang2021weakly}, more advanced consistency loss such as JS-divergence~\cite{zhang2021perturbed} and similarity weighted loss~\cite{wei2021dense}.
However, given limited annotations, exploring feature consistency only would be insufficient to capture the complex structures of point clouds, making it very difficult to extract both contextual and object information for satisfactory segmentation performance.
To inspect the consistency-based methods' comprehension of scene context, we conduct a pilot study by masked evaluation: evaluate the segmentation performance {given a \textit{context-to-be-filled} point cloud. As shown in Figure~\ref{fig:motivation}, the performance of the consistency-based method degenerates drastically, indicating a poor understanding of the scene context, even in this simple case. Thus, comprehending the complex scene context from mass unlabeled points remains an unresolved issue.


Motivated by masked modeling (\eg~MAE~\cite{he2022masked}) in image and video that learns good representations by masking random patches of the input image and reconstructing the missing information, we seek to endow the power of masked modeling for weakly-supervised point cloud segmentation. However, directly employing MAE to 3D point clouds with sparse annotations may fail to work due to the following reasons. \textbf{First}, since 3D point clouds are typically unordered and irregular, it is nontrivial to mask out the informative visual context from the 3D point clouds for subsequent context learning. \textbf{Second}, considering the limited but valuable labeled data in the weakly-annotated point cloud, how to fully exploit the labeled points in masked modeling remains an open question.

To address the above issues, we propose a simple yet effective \name (\shortname) that consists of two parts: region-wise masking (RegionMask) strategy and a contextual masked training (CMT) method.
To be specific, RegionMask evenly divides the geometric space into a set of cuboids and masks all points within the cuboids selected with a given mask ratio. Different from the trivial point-wise masking solution~\cite{min2022voxel} that performs point-wise random masking, our RegionMask masks the point cloud continuously in the geometric space to provide a meaningful masked context prediction task. Beyond that, RegionMask is able to control the difficulty of the masked feature prediction task by adjusting a hyper-parameter region size, showing flexibility in handling different amounts of annotation. {Similar to MAE~\cite{he2022masked}, we expect that with a very high mask ratio (\ie 0.75), the model is able to learn more visual concepts~\cite{he2022masked}, thereby mastering the contextual information. However, as shown in our experiments, directly incorporating the masked modeling objective into the consistency-based training framework impedes learning from the limited but valuable labeled points, resulting in degenerated performance.} To resolve this problem, we propose a contextual masked training (CMT) method that adds an extra masked feature prediction branch into the consistency-based framework, which not only paves the way for learning labeled data but allows the model to effectively learn the complex scene context. The proposed \shortname achieves state-of-the-art performance on two widely-tested benchmarks ScanNet V2 and S3DIS. For example, on ScanNet V2~\cite{dai2017scannet}, \shortname outperforms SQN~\cite{hu2022sqn} by $\best{5.6\%}$ mIoU on online test set.

Our contributions are summarized as follows:
\begin{itemize}
    
    \item We propose \lowername that incorporates masked modeling into the consistency-based training framework to effectively learn contextual information from sparsely-annotated data.

    \item We propose a region-wise masking strategy that masks the point cloud continuously to construct the meaningful masked prediction task and a contextual masked training method that facilitates the learning from limited labeled data and masked context prediction.


    \item To the best of our knowledge, we are the first to explore 3D masked modeling on weakly-supervised point cloud segmentation. {Extensive experiments on widely-tested benchmarks demonstrate the superior performance of the proposed \shortname.}
    
\end{itemize}

\begin{figure*}[!t]
    \centering
    \includegraphics[width=1.0\linewidth]{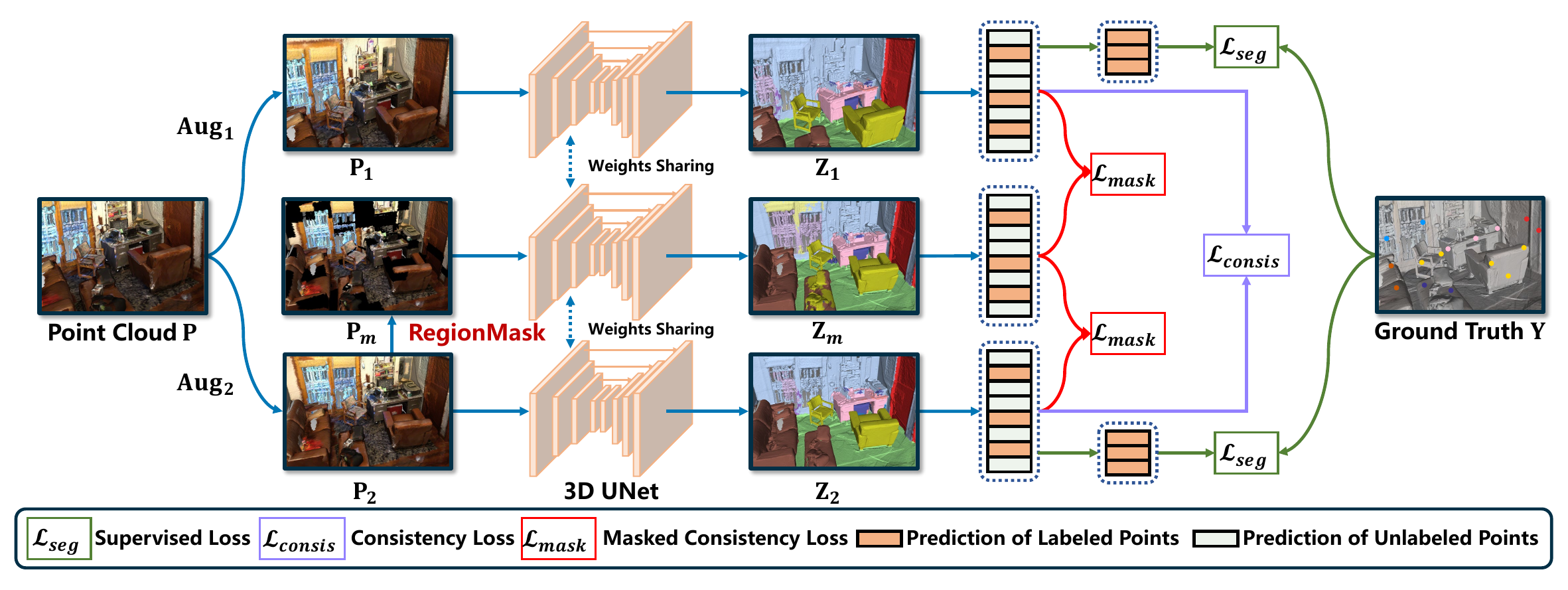}
    \caption{Overall scheme of our \shortname method. Given a point cloud $\bP$, we first apply two random augmentations and our region-wise masking to obtain the augmented point clouds $\bP_1, \bP_2$ and the masked point cloud $\bP_m$, respectively. Then, the features $\bZ_1, \bZ_2, \bZ_m$ are extracted by a weight-sharing 3D UNet. The supervised cross-entropy loss $\mL_{seg}$ is computed over labeled features and a consistency loss $\mL_{consis}$ is computed on $\bZ_1, \bZ_2$. Last, our masked consistency loss $\mL_{mask}$ enforces the feature consistency between $\bZ_1, \bZ_m$ and $\bZ_2, \bZ_m$ to help the model focus on learning contextual information.}
    \label{fig:framework_cmm}
    \vspace{-10 pt}
\end{figure*}

\section{Related Work}
\invisiblesection{Fully-supervised point cloud segmentation} 
There are mainly three kinds of fully-supervised methods proposed to encode the 3D point cloud into effective representations for semantic segmentation, including point-based~\cite{hu2020randla,landrieu2018large,li2018so,qi2017pointnet,qi2017pointnet++,wang2019exploiting,zhao2019pointweb}, voxel-based~\cite{choy20194d,graham20183d,le2018pointgrid,li2018pointcnn,liu2019densepoint,liu2019relation,rethage2018fully,riegler2017octnet,wu2019pointconv} and hybrid methods~\cite{cheng20212,xu2021rpvnet}. Early attempts~\cite{su2015multi,tatarchenko2018tangent,zhuang2021perception} simply employ the 2D convolution on the projected point cloud image, which is efficient but the projection process causes the loss of 3D geometric detail. The point-based methods are proposed to directly process the irregular and unordered points with order-agnostic architectures such as PointNet~\cite{qi2017pointnet} and PointNet++~\cite{qi2017pointnet++} that can be naturally applied to the point cloud but are less effective than 2D convolution in encoding the contextual information~\cite{hu2020randla,landrieu2018large,li2018so,wang2019exploiting,zhao2019pointweb}. The voxel-based methods~\cite{choy20194d,graham20183d,le2018pointgrid} combine the neighboring points into regular grids and often leverage sparse convolution~\cite{graham20183d,li2018pointcnn,rethage2018fully,riegler2017octnet,wu2019pointconv} to handle the sparse voxelized data. The latest works combine the merits from both worlds and form hybrid methods, but also bring more complex architecture design and extra training costs~\cite{cheng20212,xu2021rpvnet}. 
Overall, the fully-supervised point cloud segmentation methods have a strong dependence on densely-annotated labels, limiting their application scenarios.

\invisiblesection{Weakly-supervised point cloud segmentation} Learning from weakly annotated point cloud data has become a hot research topic~\cite{hu2022sqn,liu2021one,wei2020multi,wei2021dense,xu2020weakly,zhang2021weakly,zhang2021perturbed}, which not only reduces the annotation cost but also turns out to be a more general solution for real-life segmentation scenarios~\cite{liu2022less,wei2020multi}. For the partially labeled point cloud, the supervised cross-entropy loss is suitable to learn from the labeled points, which, however, is prone to learn an overfit segmentation model due to the very limited annotations~\cite{liu2021one,ren20213d,wei2021dense}. Thus, existing approaches focus on learning the major unlabeled part and can be grouped into two paradigms: pseudo labeling~\cite{hu2022sqn,liu2021one,wei2020multi} and consistency-based regularization~\cite{li2022hybridcr,xu2020weakly,yang2022mil,zhang2021weakly,zhang2021perturbed}. The pseudo-labeling methods predict pseudo-labels of the unlabeled points to explore them. MPRM~\cite{wei2020multi} trains a segmentation model on the sub-cloud labels and uses the class activation map~\cite{zhou2016learning} to pseudo-label the whole sub-cloud to train the final model. OTOC~\cite{liu2021one} improves the quality of the pseudo labels with multi rounds self-training. SQN~\cite{hu2022sqn} leverages the geometric prior to better use limited labels.
Since the pseudo label is destined to be inaccurate, consistency-based approaches learn the feature consistency across augmentations~\cite{li2022hybridcr,wei2021dense,xu2020weakly,yang2022mil,zhang2021weakly,zhang2021perturbed} or calibrated views~\cite{wei2021dense} to use mass unlabeled data. MIL~\cite{yang2022mil} enforce scene-level feature consistency for model optimization. Moreover, point-wise consistency is also leveraged by considering the color or geometric smoothness~\cite{xu2020weakly,zhang2021weakly}, feature similarity~\cite{wei2021dense,zhang2021perturbed} or using pseudo-labeling as guidance~\cite{li2022hybridcr}. However, feature consistency across augmentations may not fully comprehend the complex structures of weakly-annotated point clouds. Instead, we propose to learn masked feature consistency to better explore the contextual information.

\invisiblesection{Masked modeling for vision} Masked modeling has been a long endeavor to learn effective representation from vision data. Early attempts reconstruct RGB features from masked images~\cite{pathak2016context}, which are improved by masking a very high ratio of image content to learn meaningful visual representation~\cite{he2022masked,liu2022densely,xie2022simmim,yi2022masked}. Moreover, masked supervised learning improves the perception of contextual information in fully-supervised image semantic segmentation~\cite{zunair2022masked}. Recently, researchers apply the masked modeling approach to learn unlabeled point cloud data~\cite{liu2022masked,min2022voxel,pang2022masked}.
Unlike the above settings, weakly-supervised point cloud segmentation provides both labeled and unlabeled data. Moreover, applying masked modeling tailored for unsupervised / fully-supervised learning to both labeled and unlabeled data simultaneously is rarely explored. In this paper, we propose a contextual masked training method to learn from the limited supervision and the masked feature prediction task for weakly-supervised point cloud semantic segmentation.

\vspace{-5pt}
\section{\name}

\invisiblesection{Notations} Formally, a point cloud data is a collection of $N$ points $\bP = \{p_1, p_2, \dots, p_N\}$, where each point $p_n$ often comprises the geometric location and RGB information, \ie $p_n = \bP[n] = (x_n, y_n, z_n, r_n, g_n, b_n)$. We use $[\cdot]$ as the index operation that retrieves the corresponding element (can be a vector or a scalar) from a set or a matrix. To accomplish the point cloud semantic segmentation task, given a point cloud $\bP$ and a segmentation network $f_\theta(\cdot)$ parameterized by $\theta$, we expect the model to produce point-wise classification features\footnote{We use the term features and logits interchangeably for convenience.} $\bZ = \text{Softmax}\bigl(f_\theta(\bP)\bigr)$, where $ \bZ[n] \in (0, 1)$, $\text{argmax}\bigl(\bZ[n]\bigr) \in \mC$ and $\mC = \{0, 1, 2, \dots, C-1\}$ is a predefined category set with $C$ classes. Unlike the fully-supervised point cloud semantic segmentation that provides the label of every point in $\bP$, only sparse annotations are available in weakly-supervised point cloud semantic segmentation. The weakly-labeled point cloud data comprises two parts, the labeled part and the unlabeled part, \ie $(\bP, \bY) = \{(p_s, y_s)~|~s \in \mS \} \cup \{(p_u, \oslash)~|~u \in \mU \}$, where $\mS, \mU$ denote the index sets of the labeled and unlabeled points respectively and $\oslash$ is a special token denoting the label is unavailable. 
During model training, a dataset $\mD = \{(\bP, \bY)\}$ includes hundreds of or thousands of  point cloud \& weak-label pair is provided. 


\subsection{Problem Definition} 

With the limited labeled data and a mass of unlabeled data, weakly-supervised point cloud semantic segmentation focuses on learning useful representations from a large amount of unlabeled data to improve model generalization. Existing approaches often achieve this by enforcing point-wise feature consistency across augmentations~\cite{li2022hybridcr,xu2020weakly,yang2022mil,zhang2021perturbed}. Given a weakly-labeled point cloud data $(\bP, \bY)$, two random augmentations\footnote{\revision{Details on the data augmentation are put in the supplementary.}} are applied $\bP_1 = \text{Aug}_1(\bP)$ and $ \bP_2 = \text{Aug}_2(\bP)$. Based on this, point-wise classification for two point clouds is calculated by $ \bZ_1 = \text{Softmax}\bigl(f_\theta(\bP_1)\bigr), \bZ_2 = \text{Softmax}\bigl(f_\theta(\bP_2)\bigr)$. The general form for the consistency-based method is as follows:
\begin{equation}
    \small
    \label{eqn:consis_based}
    \mL_\text{CB} = \mL_{seg} + \alpha \mL_{consis},
\end{equation}
where $\mL_{seg}$ and $\mL_{consis}$ denote supervised cross-entropy loss and the consistency loss introduced below and $\alpha$ is a hyper-parameter that controls optimization strength on the consistency loss. The supervised loss $\mL_{seg}$ is computed over limited labeled points:
\begin{equation}
    \small
    \label{eqn:loss_seg}
    \mL_{seg} = \frac{1}{|\mS|} \sum\nolimits_{s \in \mS} {CE}\bigl( \bZ_1[s], \bY[s] \bigr) + {CE}\bigl( \bZ_2[s], \bY[s] \bigr),
\end{equation}
where ${CE}(\cdot, \cdot)$ is the cross-entropy loss. In the meanwhile, the consistency loss $\mL_{consis}$ enforces point-wise feature consistency as follows:
\begin{equation}
    \small
    \label{eqn:loss_consis}
    \mL_{consis} = \frac{1}{N} \sum\nolimits_{n}  {JS} \bigl(\bZ_1[n], \bZ_2[n] \bigr),
\end{equation}
where ${JS}(\cdot, \cdot)$ minimizes the Jensen-Shannon divergence of different features.
Feature consistency from different augmentations can exploit the unlabeled data but may not be informative enough to comprehend the complex structure of the point cloud data, failing to effectively explore the contextual information such as space, color and semantic continuity that is crucial for satisfactory segmentation. Attracted by the strong context modeling ability of masked modeling in image and video representation learning, we seek to endow the power of masked modeling to weakly-supervised point cloud segmentation. However, designing an effective masking strategy for 3D point cloud data and developing a compatible training scheme to fully exploit the limited labeled data for masked modeling remain open questions.


\begin{algorithm}[!t]
    \small
    \caption{Training method for \shortname}
    \label{alg:training_cmm}
    \begin{algorithmic}[1]
    \REQUIRE
    The training dataset $\mD = \{(\bP, \bY)\}$,
    the point cloud segmentation network $f_{\theta}(\cdot)$,
    the region size $G$, the mask ratio $R$,
    the weighting factor $\alpha, \beta$, 
    the learning rate $\eta$. \\
    \ENSURE
    Optimized point cloud segmentation network $f_{\theta}$. \\
    \STATE Randomly initializes the model parameters $\theta$. 
    
    \WHILE{not converge}
        \STATE Obtain a weakly-labeled point cloud data $(\bP, \bY)$ from $\mD$. \\
        \STATE Obtain the labeled indexes $\mS$ from $\bY$. \\
        \STATE // {\emph{{perform two random augmentations}}}
        \STATE $\bP_1 \leftarrow \text{Aug}_1(\bP), \bP_2 \leftarrow \text{Aug}_2(\bP)$. \\
        \STATE Compute region-wise masking flag $\bM$ by Eqn.~(\ref{eqn:point_mask}). \\
        \STATE Compute region-wise masked point cloud $\bP_m$ by Eqn.~(\ref{eqn:grid_mask}). \\
        \STATE // {\emph{{perform segmentation for augmented point clouds}}}
        \STATE $\bZ_1 \leftarrow \text{Softmax}\bigl(f_\theta(\bP_1)\bigr), \bZ_2 \leftarrow \text{Softmax}\bigl(f_\theta(\bP_2)\bigr)$. \\
        \STATE // {\emph{{perform segmentation for the masked point cloud}}}
        \STATE $\bZ_m \leftarrow \text{Softmax}\bigl(f_\theta(\bP_m)\bigr)$. \\
        \STATE Compute the cross-entropy loss $\mL_{seg}$ by Eqn.~(\ref{eqn:loss_seg}). \\
        \STATE Compute the consistency loss $\mL_{consis}$ by Eqn.~(\ref{eqn:loss_consis}). \\
        \STATE Compute the masked consistency loss $\mL_{mask}$ by Eqn.~(\ref{eqn:loss_masked_consistency}). \\
        \STATE Compute the overall training objective $\mL_\text{CPCM}$ by Eqn.~(\ref{eqn:cmm_training_objective}). \\
        \STATE // {\emph{update network parameters via gradient descent}}
        \STATE $\theta \leftarrow \theta - \eta \nabla_{\theta} \mL_\text{\shortname}$. \\
    \ENDWHILE
    \end{algorithmic}
\end{algorithm}


\invisiblesection{Overview} To answer the above questions, we propose \name (\shortname) to model the contextual information effectively with two steps: \textbf{First}, we propose a region-wise masking strategy that masks the point cloud in the continuous geometric space, providing meaningful missing context to be filled. \textbf{Second}, we propose a contextual masked training that facilitates the learning of limited labeled points and masked feature prediction tasks by adding an extra stream for masked feature extraction. Then, we enforce the feature consistency between masked and unmasked features to learn effective contextual representations. The overall framework and algorithm of \shortname are shown in Figure~\ref{fig:framework_cmm} and Algorithm~\ref{alg:training_cmm}, respectively.

\begin{figure}[!t]
    \centering
    \includegraphics[width=0.99\linewidth]{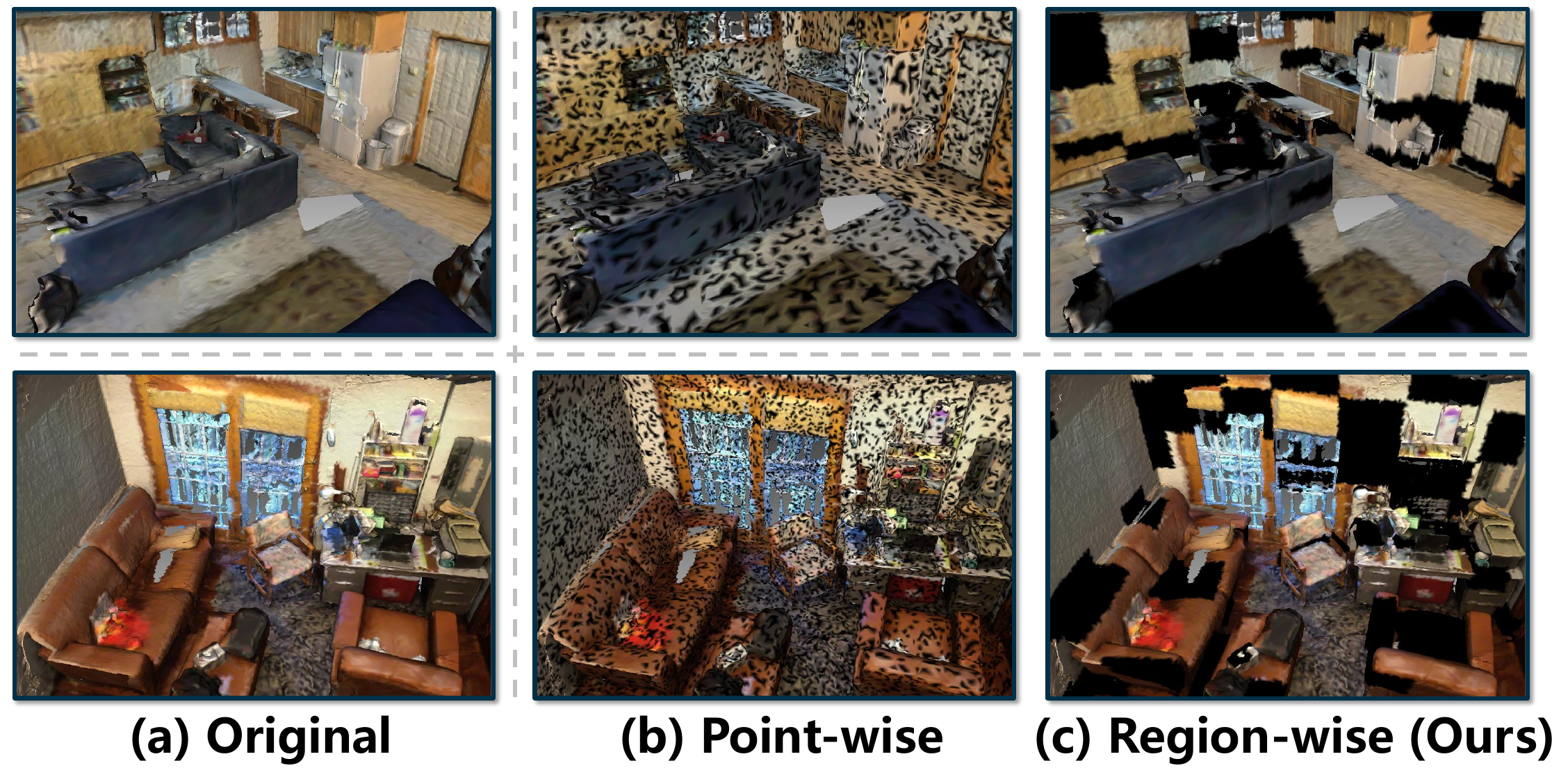}
    \caption{Comparisons of different masking strategies. The proposed region-wise masking removes meaningful context to be filled. We set the mask ratio $= 25\%$ for visualization.}
    \vspace{-15pt}
    \label{fig:grid_mask}
\end{figure}
   
\vspace{-3pt}
\subsection{Region-wise Point Cloud Masking}
In this section, we introduce our region-wise masking scheme that provides an effective supervision signal for the model to learn contextual information. To formulate the masking strategy, we first define $\bM \in \mmR^{N}$ as a zero-one vector to indicate whether a point in point cloud\footnote{For convenience, we refer to the point cloud data as a matrix.} $\bP \in \mmR^{N \times 6}$ is masked or not and denote the mask ratio as $R~(0 \le R \le 1)$, \ie the number of the masked points is $R * N$. Then, the masked point cloud $\bP_m$ is computed in a point-wise setting the color information to zero\footnote{The coordinate $x, y, z$ is left untouched since the sparse convolution operation in 3D UNet requires it for the convolution kernel construction.}:
\begin{equation}
\small
    \label{eqn:point_mask}
    \bP_m[n] = \bigl[x_n, y_n, z_n, \bM[n] \cdot r_n, \bM[n] \cdot g_n, \bM[n] \cdot b_n\bigr].
\end{equation}

To obtain a masked point cloud, a straightforward solution, termed PointMask, is to randomly sample each point (or voxel) with the given mask ratio $R$
\begin{equation}
    \small
    \bM[n] = \mathds{1}\{q \leq R \},~q \sim U[0, 1],
\end{equation}
where $\mathds{1}\{ \cdot \}$ is the indicator function and $q$ is a random variable drawn from the uniform distribution $U[0, 1]$. As shown in Table~\ref{tab:ablation_masking}, PointMask delivers unsatisfactory improvement compared to the baseline, especially with a very high mask ratio (\ie 0.75). We attribute this failure to the following reasons: The PointMask strategy tends to decrease the resolution of the point cloud (see Figure~\ref{fig:grid_mask}\textcolor{red}{b}), which does not effectively mask meaningful visual words~\cite{he2022masked} to predict. 


To reasonably remove some contextual information from a point cloud, we introduce Region-wise Masking (RegionMask) that evenly splits the scene into cuboids and masks the points within the randomly selected cuboids. We first define the region size $G$ to denote the number of cuboids. Note that a cuboid that parallels the axes in a 3D coordinate system is represented by $\bigl[(x_\text{min}, y_\text{min}, z_\text{min}), (x_\text{max}, y_\text{max}, z_\text{max})\bigr]$. Assuming that the minimal cuboid covering a point cloud is $\bigl[(0, 0, 0), (l, w, h)\bigr]$. We evenly partition the scene into a set of cuboid regions $\mH$ \ie $(|\mH| = G^3)$ as follows:
\begin{equation}
\begin{aligned}
    \footnotesize
    \label{eqn:grid_def}
    & \mH = \Bigl\{\bigl[ (x_i, y_j, z_k), (x_{i+1}, y_{j+1}, z_{k+1}) \bigr]\Bigr\}, \\
    & x_i = i \cdot \frac{l}{G}, y_j = j \cdot \frac{w}{G}, z_k = k \cdot \frac{h}{G}, \\
    & i,j,k \in \{0, 1, \dots, G - 1\},
\end{aligned}
\end{equation}
where $x_i, y_j, z_k$ are the evenly split points along the $x, y, z$ axes and  $\bigl(\frac{l}{G}, \frac{w}{G}, \frac{h}{G}\bigr)$ are the length, width, height of a region, respectively. Then, we randomly select $R \cdot G^3$ regions $\mH^m$ and compute the mask flag $\bM$ as follows:
\begin{equation}
    \small
    \label{eqn:grid_mask}
    \bM[n] = \mathds{1}\bigl\{ (x_n, y_n, z_n) \in \mH^m \bigr\},
\end{equation}
where $\in$ denotes a point that lies within a cuboid or not. Then, the masked point cloud is computed by Eqn.~(\ref{eqn:point_mask}). As shown in Figure~\ref{fig:grid_mask}\textcolor{red}{c}, RegionMask masks the unordered and irregular point cloud continuously, providing meaningful context-to-be-filled patterns such as partial inner-instance mask and cross-instance mask. Moreover, as shown in Section~\ref{sec:further_analysis}, RegionMask is able to flexibly cope with different amounts of annotation by adjusting the region size.


\vspace{-3pt}
\subsection{Contextual Masked Training Method}
\vspace{-5pt}
In this section, we introduce our contextual masked training method for learning the contextual information between the masked and unmasked data. We first consider the mask operation as a ``strong augmentation'' and incorporate it directly into the consistency-based training framework. However, as shown in Figure~\ref{fig:compare_cpcm_cmt}, the training cross-entropy error significantly increases and the performance drops considerably. These results indicate that the input distribution is significantly altered by the mask operation, which impedes learning from limited but valuable labeled points.


\begin{table*}[!h]\small
{
    \resizebox{0.999\linewidth}{!}{
    \begin{tabular}{r|r|c|ccccccccccccc}
    \hline
    Method & Setting & mIoU (\%) & \rotateboxtable{ceiling} & \rotateboxtable{floor} & \rotateboxtable{wall} & \rotateboxtable{beam} & \rotateboxtable{column} & \rotateboxtable{window} & \rotateboxtable{door} & \rotateboxtable{chair} & \rotateboxtable{table} & \rotateboxtable{bookcase} & \rotateboxtable{sofa} & \rotateboxtable{board} & \rotateboxtable{clutter} \\
    \hline\hline
    MinkNet$^*$~\cite{choy20194d}& \multirow{5}{*}{Fully} & 68.2 & 91.7 & 98.7 & 83.8 & 0.0 & 24.7 & 56.8 & 72.1 & 91.5 & 83.5 & 73.3 & 70.8 & 81.3 & 58.4   \\
    PointNet~\cite{qi2017pointnet} &  & 41.1 & 88.8 & 97.3 & 69.8 & 0.1 & 4.0 & 46.3 & 10.8 & 58.9 & 52.6 & 5.9 & 40.3 & 26.4 & 33.2 \\
    KPConv~\cite{thomas2019kpconv} &  & 67.1 & 92.8 & 97.3 & 82.4 & 0.0 & 23.9 & 58.0 & 69.0 & 91.0 & 81.5 & 75.3 & 75.4 & 66.7 & 58.9 \\
    RandLA-Net~\cite{hu2020randla} &  & 62.4 & 91.2 & 95.7 & 80.1 & 0.0 & 25.2 & 62.3 & 47.4 & 75.8 & 83.2 & 60.8 & 70.8 & 65.2 & 54.0 \\
    RFCR~\cite{gong2021omni} &  & 68.7 & 94.2 & 98.3 & 84.3 & 0.0 & 28.5 & 62.4 & 71.2 & 92.0 & 82.6 & 76.1 & 71.1 & 71.6 & 61.3\\
    \hline\hline
    $\Pi$ Model~\cite{laine2016temporal} & \multirow{3}{*}{10\%} & 46.3 & 91.8 & 97.1 & 73.8 & 0.0 & 5.1 & 42.0 & 19.6 & 66.7 & 67.2 & 19.1 & 47.9 & 30.6 & 41.3 \\
    MT~\cite{tarvainen2017weight} &  & 47.9 & 92.2 & 96.8 & 74.1 & 0.0 & 10.4 & 46.2 & 17.7 & 67.0 & 70.7 & 24.4 & 50.2 & 30.7 & 42.2 \\
    10$\times$Fewer~\cite{xu2020weakly} &  & 48.0 & 90.9 & 97.3 & 74.8 & 0.0 & 8.4 & 49.3 & 27.3 & 69.0 & 71.7 & 16.5 & 53.2 & 23.3 & 42.8 \\
    \arrayrulecolor{lightgray}\hdashline
    SPT~\cite{zhang2021weakly} & \multirow{3}{*}{1\%} & 61.8 & 91.5 & 96.9 & 80.6 & 0.0 & 18.2 & 58.1 & 47.2 & 75.8 & 85.7 & 65.3 & 68.9 & 65.0 & 50.2 \\
    PSD~\cite{zhang2021perturbed} &  & 63.5 & 92.3 & 97.7 & 80.7 & 0.0 & 27.8 & 56.2 & 62.5 & 78.7 & 84.1 & 63.1 & 70.4 & 58.9 & 53.2 \\
    HybridCR~\cite{li2022hybridcr} &  & 65.3 & 92.5 & 93.9 & 82.6 & 0.0 & 24.2 & 64.4 & 63.2 & 78.3 & 81.7 & 69.0 & 74.4 & 68.2 & 56.5 \\
    \arrayrulecolor{lightgray}\hdashline
    $\Pi$ Model~\cite{laine2016temporal} & \multirow{3}{*}{0.2\%} & 44.3 & 89.1 & 97.0 & 71.5 & 0.0 & 3.6 & 43.2 & 27.4 & 62.1 & 63.1 & 14.7 & 43.7 & 24.0 & 36.7 \\
    MT~\cite{tarvainen2017weight} &  & 44.4 & 88.9 & 96.8 & 70.1 & {0.1} & 3.0 & 44.3 & 28.8 & 63.6 & 63.7 & 15.5 & 43.7 & 23.0 & 35.8 \\
    10$\times$Fewer~\cite{xu2020weakly} &  & 44.5 & 90.1 & {97.1} & 71.9 & 0.0 & 1.9 & 47.2 & 29.3 & 62.9 & 64.0 & 15.9 & 42.2 & 18.9 & 37.5 \\
    \arrayrulecolor{lightgray}\hdashline
    SQN~\cite{hu2022sqn} & \multirow{2}{*}{0.1\%} & 61.4 & \best{91.7} & \best{95.6} & 78.7 & 0.0 & 24.2 & 55.9 & 63.1 & 62.9 & 70.5 & 67.8 & 60.7 & 56.1 & 50.6 \\
    \shortname (Ours) &  & ~~~~~~~\bestimprove{66.3}{4.9} & 91.4 & 95.5 & \best{82.0} & 0.0 & \best{30.8} & \best{54.1} & \best{70.1} & \best{87.6} & \best{79.4} & \best{70.0} & \best{67.0} & \best{77.8} & \best{56.6} \\
    \arrayrulecolor{black}
    \hline
    PSD~\cite{zhang2021perturbed} & \multirow{2}{*}{0.03\%} & 48.2 & 87.9 & 96.0 & 62.1 & 0.0 & 20.6 & 49.3 & 40.9 & 55.1 & 61.9 & 43.9 & 50.7 & 27.3 & 31.1 \\
    HybridCR~\cite{li2022hybridcr} &  & 51.5 & 85.4 & 91.9 & 65.9 & 0.0 & 18.0 & 51.4 & 34.2 & 63.8 & 78.3 & 52.4 & 59.6 & 29.9 & 39.0 \\
    \arrayrulecolor{lightgray}\hdashline
    MIL~\cite{yang2022mil} & \multirow{3}{*}{0.02\%} & 51.4 & \na & \na & \na & \na & \na & \na & \na & \na & \na & \na & \na & \na & \na \\
    MIL$^*$~\cite{yang2022mil} &  & 52.1 & 89.2 & 95.5 & 74.8 & \best{0.2} & \best{19.2} & 41.1 & 23.1 & 76.3 & 64.7 & 62.6 & 27.8 & 57.8 & 44.8 \\
    \shortname (Ours) & & ~~~~~~~~~\bestimprove{62.3}{10.2} & \best{92.6} & \best{95.6} & \best{79.4} & 0.0 & 17.8 & \best{49.3} & \best{59.4} & \best{85.7} & \best{75.6} & \best{69.1} & \best{60.7} & \best{68.2} & \best{55.8} \\
    \arrayrulecolor{black}
    \hline
    \end{tabular}
    }
}
\centering
\caption{Comparisons with state-of-the-art methods on S3DIS area5 test set. $*$ denotes results based on our reimplementation.
}
\vspace{-15pt}
\label{tab:sota_per_class_s3dis}
\end{table*}

\invisiblesection{Training objective} Taking both the learning from limited labeled data and the learning of contextual information into account, we propose to add an extra branch to perform the masked features prediction task while leaving the two weakly-supervised branches untouched. To be specific, given a weakly-labeled point cloud data $(\bP, \bY)$, we obtain two point clouds $\bP_1, \bP_2$ by two random augmentations and the masked version $\bP_m$ by the proposed RegionMask. Then, we extract their corresponding features $\bZ_1, \bZ_2, \bZ_m$ with the segmentation model $\text{Softmax}\bigl( f_{\theta}(\cdot) \bigr)$. Last, the overall training objective for our contextual masked training is as follows
\begin{equation}
    \label{eqn:cmm_training_objective}
    \mL_\text{CPCM} = \mL_{seg} + \alpha \mL_{consis} + \beta \mL_{mask},
\end{equation}
where $\beta$ is a hyper-parameter to control the optimization strength of contextual masked learning and $\mL_{mask}$ is our masked consistency loss introduced below.

\invisiblesection{Masked consistency loss} We seek to learn contextual information through masked and unmasked features. To this end, we propose to minimize the distribution gap between masked and unmasked features. In this way, the model shall learn to leverage the unmasked part in the masked point cloud \ie the surrounding context, thereby improving segmentation performance. Specifically, with the features $\bZ_1, \bZ_2, \bZ_m$ respectively extracted from the two randomly augmented and the masked point clouds, we introduce our masked consistency loss as follows:
\begin{equation}
    \small
    \label{eqn:loss_masked_consistency}
    \mL_{mask} = \frac{1}{N} \sum_{n}{JS}\bigl(\bZ_1[n], \bZ_m[n]\bigr) + {JS}\bigl(\bZ_2[n], \bZ_m[n]\bigr),
\end{equation} 
where the unmasked features $\bZ_1, \bZ_2$ are considered as the ``ground truth'' and we detach the gradients of $\bZ_1, \bZ_2$ during masked consistency loss calculation.
\vspace{-8pt}
\section{Experiments}
\vspace{-5pt}
\invisiblesection{Datasets} We consider two benchmark datasets ScanNet V2~\cite{dai2017scannet} and S3DIS~\cite{armeni20163d}. ScanNet V2 has 20 semantic classes and the number of training~/~validation~/~testing scans is 1,201~/~312~/~100 respectively. We evaluate our model on both val and online test set following~\cite{hu2022sqn,li2022hybridcr,yang2022mil}. {S3DIS, a large-scale point cloud dataset, contains 6 areas with 271 rooms and 13 semantic categories. We adopt the widely-used area5 test set~\cite{xu2020weakly,zhang2021perturbed} for evaluation, where the number of training and testing scans is 204 and 68, respectively.}

\invisiblesection{Implementation details} We implement our method using {MinkowskiEngine}~\cite{choy20194d}, a sparse convolution library based on PyTorch~\cite{paszke2019pytorch}, as done in previous works~\cite{hu2022sqn,yang2022mil}. As for the model architecture, we adopt the 34-layer Sparse Residual U-Net~\cite{ronneberger2015u} following previous works~\cite{hou2021exploring,xie2020pointcontrast}. For evaluation, we use the class-wise Intersection over Union (IoU) and mean IoU (mIoU) metrics. For optimization, we employ the SGD optimizer with $\text{lr} = 1e^{-2}$, weight decay = $1e^{-3}$, the polynomial learning rate scheduler with $\text{decay rate} = 0.9$ and set the batch size to 2 and 4 for ScanNet V2 and S3DIS, respectively. During training, the voxel size is set to 2cm and 5cm for ScanNet V2 and S3DIS, respectively. All models are trained for 180 epochs. We choose JS-divergence as our consistency loss~\cite{zhang2021perturbed}. We refer to the annotation ratio $< 0.1\%$ (including 20 points on ScanNet V2) as the extreme-limited annotations and $\geq 0.1\%$ as the limited annotations. As for the region size $G$ and mask ratio $R$ in RegionMask, we set the mask ratio $R = 0.75$ and set $G = 8$ and $= 4$ for the extreme-limited and limited annotations, respectively. As for $(\alpha, \beta)$ in $\mL_\text{CPCM}$, we set $(\alpha, \beta) = (5, 10)$ and $ = (1, 5)$ for the extreme-limited and limited annotations, respectively.\footnote{Analysis on hyper-parameters $\alpha, \beta$ are put in the supplementary.} All experiments are conducted on 2 and 1 TITAN 3090 GPU(s) for ScanNet V2 and S3DIS, respectively. \revision{Our source code is publicly available at \url{https://github.com/lizhaoliu-Lec/CPCM}.}

\subsection{Comparison with State-of-the-arts}

\begin{table}[!th]
{
\resizebox{0.75\linewidth}{!}{\small
    \begin{tabular}{r|r|ll}
    \hline
    Method & Setting & Val & Test \\
    \hline\hline
    PointNet++~\cite{qi2017pointnet++} & \multirow{3}{*}{Fully} & \na  & 33.9 \\
    KPConv~\cite{thomas2019kpconv} &  & \na & 68.4 \\
    MinkNet~\cite{choy20194d} &  & 72.9 & 73.6 \\
    \hline\hline 
    MPRM~\cite{wei2020multi} & \multirow{3}{*}{Scene} & 21.9 & \na \\
    WYPR~\cite{ren20213d} &  & 29.6 & 24.0 \\
    MIL~\cite{yang2022mil} &   & 26.2 & \na \\
    \hline
    MPRM~\cite{wei2020multi} & \multirow{2}{*}{Subcloud} & 43.2 & 41.1 \\
    MIL~\cite{yang2022mil} &  & 47.4 & 45.8 \\
    \hline
    SPT~\cite{zhang2021weakly} & \multirow{3}{*}{1\%} & \na & 51.1 \\
    PSD~\cite{zhang2021perturbed} &  & \na & 54.7 \\
    HybridCR~\cite{li2022hybridcr} &  & 56.9 & 56.8 \\
    \arrayrulecolor{lightgray}\hdashline
    SQN~\cite{hu2022sqn} & \multirow{2}{*}{0.1\%} & 58.4 & 56.9 \\
    \shortname (Ours) &  & \bestimprove{63.8}{5.4} & \bestimprove{62.5}{5.6} \\
    \arrayrulecolor{black}
    \hline
    WYPR~\cite{ren20213d} & \multirow{4}{*}{20 pts} & 51.5 & \na \\
    OTOC$^\dag$~\cite{liu2021one} &  & 55.1 & \na \\
    MIL~\cite{yang2022mil} &  & 57.8 & 54.4 \\
    \shortname (Ours) &  & \bestimprove{62.7}{4.9} & \bestimprove{62.8}{8.4} \\
    \hline
    \end{tabular}
}
}
\centering
\caption{Comparisons with state-of-the-art methods on ScanNet V2. $\dag$ indicates results reproduced by MIL~\cite{yang2022mil}.
}
\vspace{-10pt}
\label{tab:sota_scannet}
\end{table}

\vspace{-5pt}
\invisiblesection{Quantitative results on S3DIS}
{We provide the quantitative results on S3DIS in Table~\ref{tab:sota_per_class_s3dis}. For fair comparisons, our approach is evaluated under the same settings used by prior works \ie the annotation ratio being 0.2\%, 0.1\%, and 0.02\%. The proposed \shortname consistently outperforms the previous state-of-the-art across different annotation ratios, often by a large margin. To be specific, \shortname outperforms SQN by $4.9\%$ under the 0.1\% setting and beats MIL by $10.2\%$ under the extreme-limited annotation setting 0.02\%. Notably, our \shortname trained by 0.1\% label is able to surpass the HybridCR trained by 1\% label. By diving into per-class mIoU, we observe that our \shortname performs well in relatively small instance categories in a scene such as ``chair'', ``table'', and ``sofa'' that tend to be misclassified, which cannot be accomplished without effectively understanding the scene context.
Moreover, with $0.1\%$ annotations only, \shortname achieves competitive performance to the fully supervised MinkNet ($66.3~\vs 68.2$), closing the gap between fully and weakly supervised methods.}


\invisiblesection{Quantitative results on ScanNet V2}
 {We evaluate our approach under 0.1\% and 20 points (pts) settings on ScanNet V2 and the quantitative results are shown in Table~\ref{tab:sota_scannet}. Although the amount of annotation is very limited, the proposed \shortname provides substantial improvements over prior SoTAs. Specifically, on the validation set, \shortname leads SQN by $5.4\%$ under the 0.1\% setting and MIL by $4.9\%$ under the 20 pts setting. Moreover, on the private test set, \shortname still leads SQN and MIL by $5.6\%$ and $8.4\%$ respectively, showing the strong generalization ability of \shortname.}

\vspace{-3pt}
\subsection{Ablation Analysis on \shortname}
\vspace{-5pt}
\invisiblesection{Comparisons to baselines}
Since our implementation is based on the fully-supervised MinkNet and the weakly-supervised consis-based method, we directly compare them to investigate the effectiveness of \shortname. The results are shown in Table~\ref{tab:ablation_two_baselines}. {MinkNet performs decently with 0.1\% annotation ratio but suffers from extreme-limited annotation 0.01\%. The consis-based method delivers noticeable improvements on both datasets for all settings, showing that it is a strong baseline. Unsurprisingly, the proposed \shortname completely beats the MinkNet and the consis-based baseline, often by a large margin. Notably, when it comes to the extreme-limited 0.01\% setting, \shortname boosts the performance of MinkNet by   $14.6\%$ and $11.6\%$ on ScanNet V2 and S3DIS, respectively. These results demonstrate the advantage of \shortname that effectively comprehends the scene context over the strong consis-based baseline.} 


\begin{table}[!t]
{
\resizebox{\linewidth}{!}{
    \begin{tabular}{r|cc|ll|ll}
    \hline
    \multirow{2}{*}{Method} & \revision{\multirow{2}{*}{$\mL_{consis}$}} & \revision{\multirow{2}{*}{$\mL_{mask}$}} & \multicolumn{2}{c|}{ScanNet V2} & \multicolumn{2}{c}{S3DIS}  \\
    \cline{4-7}
     & & & $0.01\%$ & $0.1\%$ & $0.01\%$ & $0.1\%$ \\
    \hline\hline
    MinkNet & \without & \without & 37.6 & 60.3 & 47.7 & 62.9 \\ 
    \hline
    Consis-based & \with & \without & \secondimprove{44.2}{6.6} & \secondimprove{61.8}{1.5} & \secondimprove{52.9}{5.2} & \secondimprove{64.9}{2.0} \\
    \shortname (Ours) & \with & \with & \bestimprove{52.2}{14.6} & \bestimprove{63.8}{3.5} & \bestimprove{59.3}{11.6} & \bestimprove{66.3}{3.4} \\
    \hline
    \end{tabular}
}
}
\centering
\caption{Comparisons with two strong baselines: \textit{fully-supervised} method MinkNet trained on weakly-annotated labels and the \textit{weakly-supervised} consis-based method.
}
\vspace{-10pt}
\label{tab:ablation_two_baselines}
\end{table}

\begin{table}[!h]
{
\resizebox{0.999\linewidth}{!}{\small
    \begin{tabular}{l|ll|ll}
    \hline
    \multirow{2}{*}{Masking Strategy} & \multicolumn{2}{c|}{ScanNet V2 (0.01\%)} & \multicolumn{2}{c}{S3DIS (0.01\%)} \\
    \cline{2-5}
     & 0.15 & 0.75 & 0.15 & 0.75 \\
    \hline\hline
    \revision{Consis-based} & \multicolumn{2}{c|}{44.2} & \multicolumn{2}{c}{52.9} \\
    \hline
    PointMask & \noimprove{42.3}{1.9} & \secondimprove{48.2}{4.0} & \noimprove{52.3}{0.6} & \secondimprove{55.1}{2.2} \\
    RegionMask (Ours) & \bestimprove{46.5}{2.3} & \bestimprove{52.2}{8.0} & \bestimprove{55.8}{2.9} & \bestimprove{59.3}{6.4} \\
    \hline
    \end{tabular}
}
}
\centering
\caption{Ablation studies on different masking strategies. The contextual masked training modeling scheme is employed. Otherwise, all masking strategies show degenerated performance compared to the consis-based baseline.}
\label{tab:ablation_masking}
\end{table}

\vspace{-5pt}
\invisiblesection{Region masking} Since random point masking is a common solution in masked vision modeling and has recently been applied to unsupervised point cloud data learning~\cite{min2022voxel}. We investigate the behavior of PointMask under both low and high mask ratios and the results are put in Table~\ref{tab:ablation_masking}. On one hand, when the mask ratio is low (0.15), PointMask performs even slightly worse than the consis-based baseline while the proposed RegionMask boosts the performance by $2.3\%$ and $2.9\%$ on the ScanNet V2 and S3DIS, respectively. On the other hand, when the mask ratio is high (0.75), RegionMask considerably improves the performance while PointMask brings only a relatively marginal boost. We conclude that RegionMask is able to mask more meaningful visual words than PointMask under both low and high mask ratios, paving the path of promising masked vision modeling for weakly-supervised point cloud segmentation.


    


\begin{figure}[!h]
    \centering
    \begin{minipage}{\linewidth}
        \centering
        \includegraphics[width=.49\textwidth]{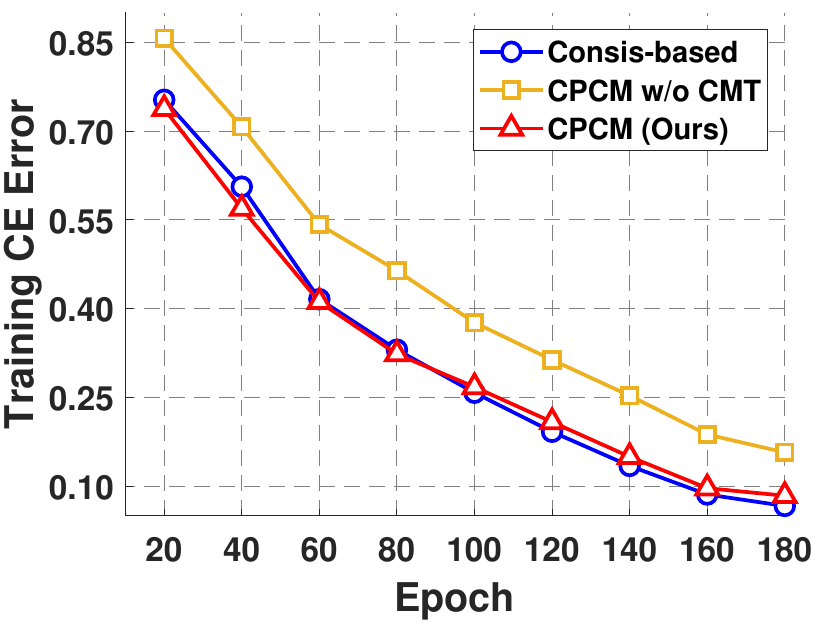}
        \includegraphics[width=.49\textwidth]{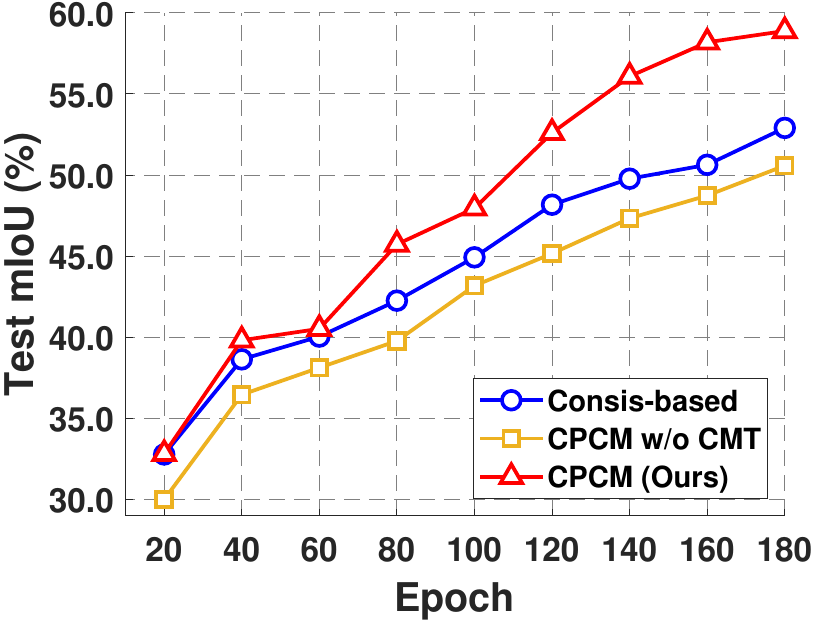}
    \end{minipage}
    
    \caption{Evolution of training cross-entropy (CE) error and test mIoU \wrt training epochs on S3DIS (0.01\%).}
    \label{fig:compare_cpcm_cmt}
\end{figure}

\vspace{-5pt}
\invisiblesection{Contextual masked training} We investigate the effectiveness of the proposed contextual masked training (CMT) by removing the masking stream, resulting in a consistency-based framework with a ``masking augmentation''. As shown in Figure~\ref{fig:compare_cpcm_cmt}, the training cross-entropy error drastically increases without CMT, which indicates simply incorporating ``masking augmentation'' hampers the learning of limited but valuable labeled data. With CMT, the segmentation model shows low training cross-entropy error as well as high test mIoU. Moreover, we also put the quantitative results in Table~\ref{tab:ablation_cmt_mcl} and observe a noticeable performance drop when discarding CMT. Then, with CMT, \shortname achieves substantial improvements over the consis-based baseline. These results verify that \shortname facilitates the learning of valuable annotation but also rich context information, achieving substantial improvements.

\begin{table}[!h]
{
\resizebox{0.999\linewidth}{!}{\small
    \begin{tabular}{cc|ll|ll}
    \hline
    \multirow{2}{*}{RM} & \multirow{2}{*}{CMT}  & \multicolumn{2}{c|}{ScanNet V2} & \multicolumn{2}{c}{S3DIS} \\
    \cline{3-6}
    & & 0.01\% & 0.1\% & 0.01\% & 0.1\% \\
    \hline\hline
    \without & \without & 44.2 & 61.8 & 52.9 & 64.9 \\
    \hline
    \with & \without & \noimprove{41.6}{2.6} & \noimprove{58.6}{3.2} & \noimprove{51.1}{1.8} & \noimprove{63.6}{1.3} \\
    \with & \with & \bestimprove{52.2}{6.7} & \bestimprove{63.8}{2.0} & \bestimprove{59.3}{6.4} & \bestimprove{66.3}{1.4}\\
    \hline
    \end{tabular}
}
}

\centering
\vspace{-5pt}
\caption{Ablation studies on our contextual masked training scheme. RM and CMT are short for RegionMask strategy and contextual masked training, respectively.}

\label{tab:ablation_cmt_mcl}
\end{table}
\vspace{-10pt}


 \begin{figure}[!h]
    \centering
    \begin{minipage}{\linewidth}
        \centering
        \begin{subfigure}{0.49\textwidth}
            \includegraphics[width=\linewidth]{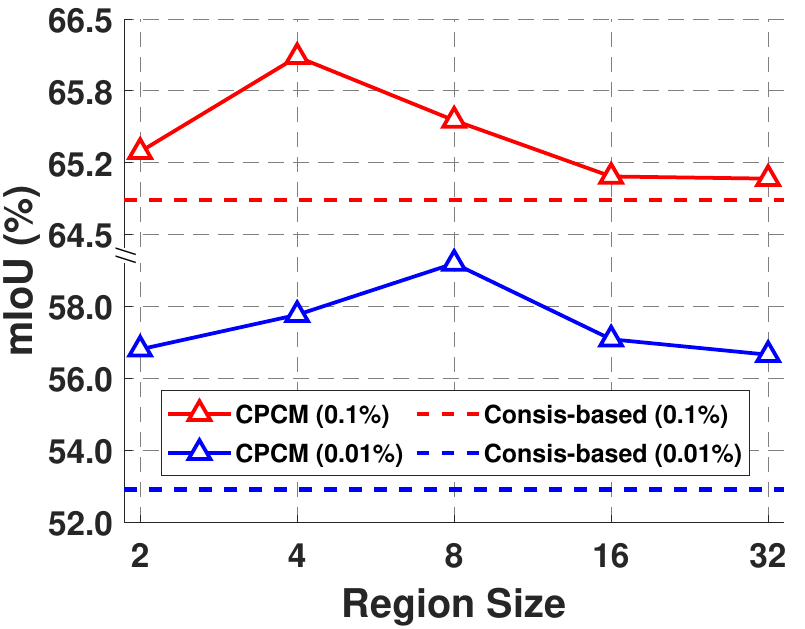}
            \caption{Effect of region size.}
            \label{fig:region_size_compare}
        \end{subfigure}
        \centering
        \begin{subfigure}{0.49\textwidth}
            \includegraphics[width=\linewidth]{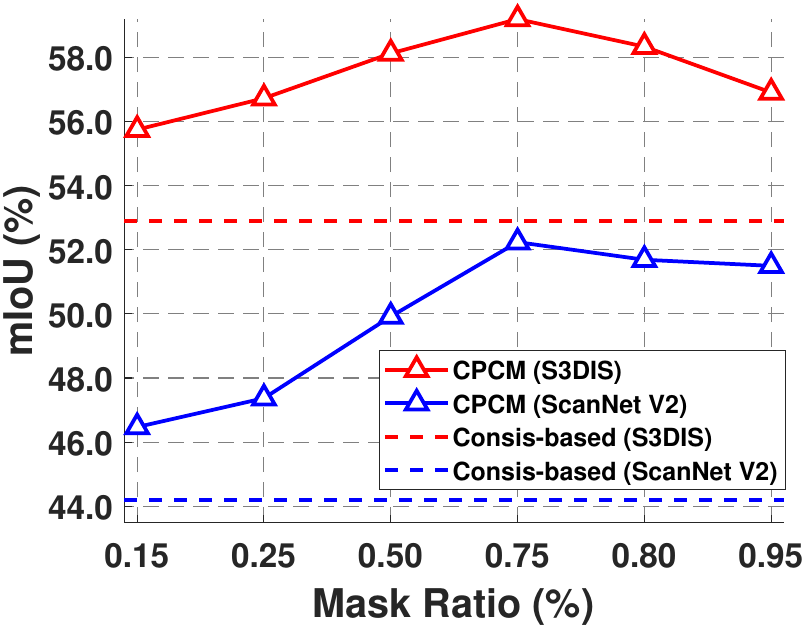}
            \caption{Effect of mask ratio.}
            \label{fig:mask_ratio_compare}
        \end{subfigure}
    \end{minipage}
    \vspace{-5pt}
    \caption{Further analysis on the proposed \shortname. (a) We investigate the effect of region size on S3DIS under 0.01\% and 0.1\% settings. (b) We investigate the effect of mask ratio on S3DIS and ScanNet V2 under the 0.01\% setting.}
    \label{fig:further_analysis}
\end{figure}

\begin{figure}[!t]
    \centering
    \includegraphics[width=\linewidth]{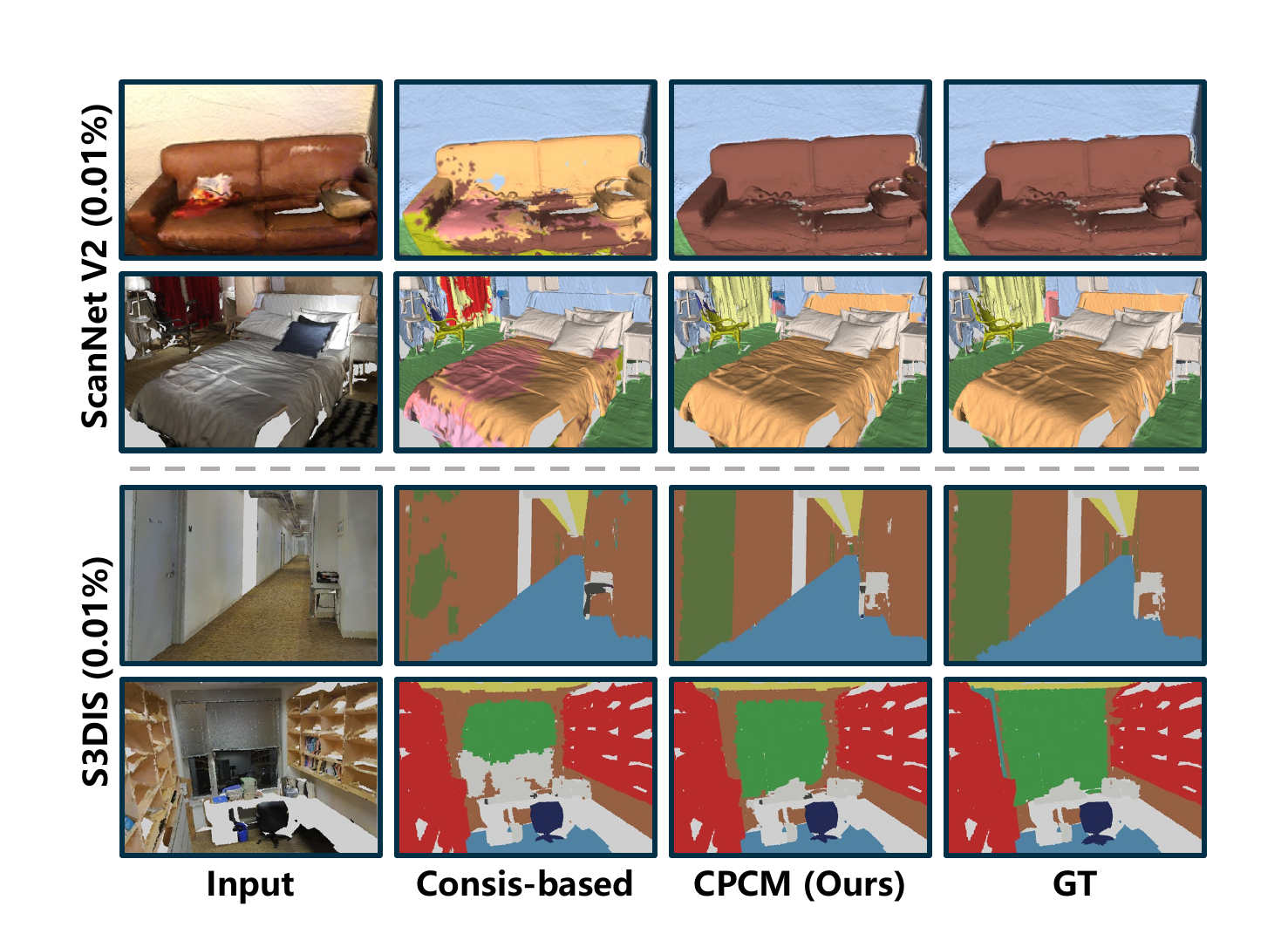}
    \caption{Qualitative comparison between the consis-based method and our \shortname on the ScanNet V2 and S3DIS.\protect\footnotemark}
    \vspace{-15pt}
    \label{fig:vis_compare}
\end{figure}

\footnotetext{More qualitative results can be found in the supplementary.}
\vspace{-15pt}

\subsection{Further Analysis on \shortname}

\label{sec:further_analysis}
\invisiblesection{Region size} As the region size increases, the task of contextual information comprehension becomes easier since the masked region to predict becomes smaller. Therefore, we are able to control the difficulty of the context comprehension task by varying the region size. With less annotation, we may set the masked features prediction task easier. In Figure~\ref{fig:region_size_compare}, the optimal region size becomes smaller when the annotation ratio goes up \ie 8 for $0.01\%$ and 4 for $0.1\%$, which verifies the flexibility of the proposed RegionMask strategy for handling different annotation ratios.

\invisiblesection{Mask ratio} More meaningful visual context will be covered as the mask ratio grows. As shown in Figure~\ref{fig:mask_ratio_compare}, the segmentation performance is constantly boosted by a larger mask ratio up to $0.75$, showing the strong potential of our \shortname to effectively explore the scene context. The optimal mask ratio is $0.75$ and exceeds which the masked context prediction task becomes too hard to achieve the best result.

\invisiblesection{Qualitative results} To intuitively understand our \shortname's ability to effectively comprehend contextual information, we provide visual comparison results in Figure~\ref{fig:vis_compare}. We first observe that \shortname shows advantages in understanding semantic categories with diverse appearances (sofa, row 1) and covering geometrically large objects (curtain and bed, row 2). Moreover, we recognize that \shortname does an excellent job at distinguishing both geometric and appearance similar categories (door and wall, row 3) and objects with complex structures (window, row 4). 
\vspace{-8pt}
\section{Conclusion}
\vspace{-5pt}
In this work, we study the learning of contextual information in the weakly-supervised point-cloud segmentation task which is not well-explored by existing methods. To this end, we proposed \shortname to model the contextual relationship among mass unlabeled points by enforcing the masked feature consistency. We first introduce a region-wise masking strategy to effectively and flexibly mask the point cloud to produce context-to-be-filled data for subsequent learning. Then, we proposed a contextual masked training method to help the model capture contextual information from both limited labeled data and the masked features prediction task. Extensive experiments on the weakly-supervised point cloud segmentation benchmarks show the superior performance of our method.
In the future, we will further explore the masked modeling scheme in the weakly-supervised point cloud detection and instance segmentation.

\noindent\textbf{Acknowledgements.} \revision{This work was partially supported by Key-Area Research and Development Program of Guangdong Province 2019B010155001, National Natural Science Foundation of China (NSFC) (62072190), National Natural Science Foundation of China (NSFC) 61836003 (key project), Program for Guangdong Introducing Innovative and Entrepreneurial Teams 2017ZT07X183.}

{\small
\bibliographystyle{ieee_fullname}
\bibliography{main_arxiv}
}
\clearpage


\section*{Supplementary material}

# CPCM: Contextual Point Cloud Modeling for Weakly-supervised Point Cloud Semantic Segmentation (Supplementary Materials)

Lizhao Liu[1,2]  Zhuangwei Zhuang[1,2]  Shangxin Huang[1]  Xunlong Xiao[1]  Tianhang Xiang[1]
Cen Chen[1]  Jingdong Wang[3]  Mingkui Tan[1,2†]
[1]South China University of Technology  [2]Pazhou Lab  [3]Baidu Inc.
{selizhaoliu, z.zhuangwei, sevtars, sexxl, sexiangtianhang}@mail.scut.edu.cn,
{chencen, mingkuitan}@scut.edu.cn, wangjingdong@baidu.com

We organize our supplementary materials as follows:

- In Section A, we provide more experiment details and results on pilot studies that inspect the contextual comprehension ability learned by the consis-based baseline and the proposed Contextual Point Cloud Modeling (CPCM) method.

- In Section B, we introduce the details of three losses, namely, the supervised cross-entropy loss $\mathcal{L}_{seg}$, the consistency loss $\mathcal{L}_{consis}$ and the proposed masked consistency loss $\mathcal{L}_{mask}$, in the objective of CPCM.

- In Section C, we present the technical details of data augmentation for the point cloud data.

- In Section D, we study the effect of hyper-parameters $\alpha$ and $\beta$ that control the optimization strength on the consistency loss and the mask consistency loss, respectively.

- In Section E, we conduct ablation studies on the masked consistency loss $\mathcal{L}_{mask}$.

- In Section F, we supply experiments on the outdoor dataset SemanticKITTI [2].

- In Section G, we conduct further experiments on the transformer architecture PTv2 [6].

- In Section H, we compare the performance of the proposed CPCM with unsupervised pre-training methods.

- In Section I, we give more implementation details on producing a strong consistency-based baseline.

- In Section J, we provide more visualization results on ScanNet V2 and S3DIS.

- In Section K, we analyze the failure case of the proposed CPCM.

## A. More Results on Pilot Studies

In this section, we inspect the context comprehension ability learned by the consis-based baseline and the proposed CPCM by a series of pilot studies. To this end, we conduct experiments with the model trained by the proposed CPCM and the consis-based training methods [5, 9]. To be specific, we train the model on two datasets: ScanNet V2 [3] with 0.01% annotations and S3DIS [1] with 0.01% annotations. Then, we design a masked evaluation protocol introduced below to quantitatively and qualitatively analyze each model's context comprehension ability.

**Masked evaluation**. We require the model to perform segmentation given a masked point cloud as input. In this sense, the masked part serves as *context-to-be-filled* and the model shall understand the masked parts' surroundings, aka contextual information, for accurate segmentation. The masked point cloud is obtained by three masking strategies detailed below.

**Masking strategies**. We introduce three masking strategies: partial-instance masking and complete instance masking, which evaluate the context comprehension within the instance and region-wise masking, which evaluates the context understanding across instances. To be specific, **1) Partial-instance masking** randomly masks some RGB features within each instance. **2) Region-wise masking** divides the point cloud into a set of regions and masks all RGB features of the randomly selected regions. **3) Complete-instance masking** completely erases the RGB features of the randomly selected instances. *Note that we use the instance annotation on ScanNet V2 and S3DIS for masked evaluation only and no instance annotation is used for model training.* During the evaluation, for three kinds of masking strategies, we gradually increase the mask ratio to increase the difficulty of the context comprehension task.

**Results**. The mIoU results w.r.t. different mask ratios are shown in Figures I and III for ScanNet V2 and S3DIS, respectively. As the mask ratio increases, we observe that our CPCM slightly decreases $0.08\% \sim 0.21\%$ on ScanNet V2 and $4.11\% \sim 5.15\%$ on S3DIS, while the consis-based baseline considerably drops $3.44\% \sim 4.14\%$ on ScanNet V2 and $10.37\% \sim 12.26\%$ on S3DIS. We also present the visual comparison results in Figures II and IV for ScanNet V2 and S3DIS, respectively. Note that we select the visual results under mask ratio $40\%$ for better visualization. We can see that our CPCM performs well under both standard and masked evaluations while the consis-based baseline fails to fill the masked part. These results demonstrate that the proposed CPCM has much stronger context comprehension ability over the consis-based baseline and achieves better and more robust point cloud semantic segmentation.

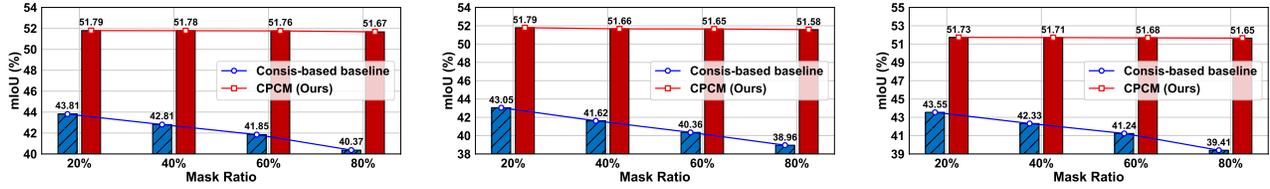

(a) Partial-instance masked evaluation results.  (b) Region-wise masked evaluation results.  (c) Complete-instance masked evaluation results.

Figure I: Masked evaluation results on ScanNet V2 [3] to inspect the contextual perception ability.

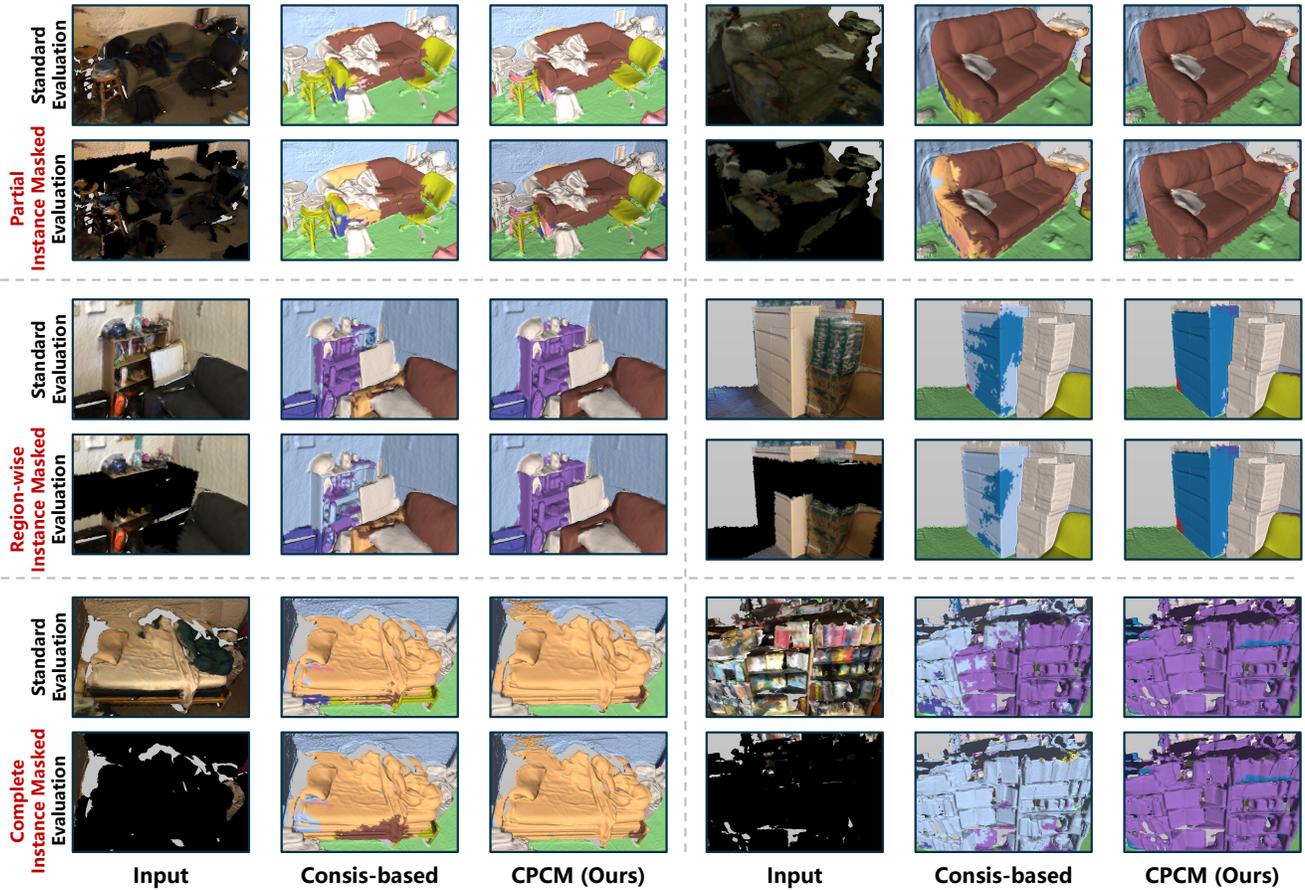

Figure II: Visual comparison of results from different methods on ScanNet V2 (mask ratio is 40%).

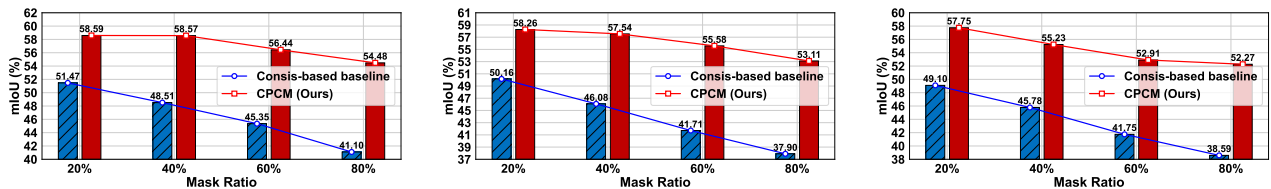

(a) Partial-instance masked evaluation results.    (b) Region-wise masked evaluation results.    (c) Complete-instance masked evaluation results.

Figure III: Masked evaluation results on S3DIS [1] to inspect the contextual perception ability.

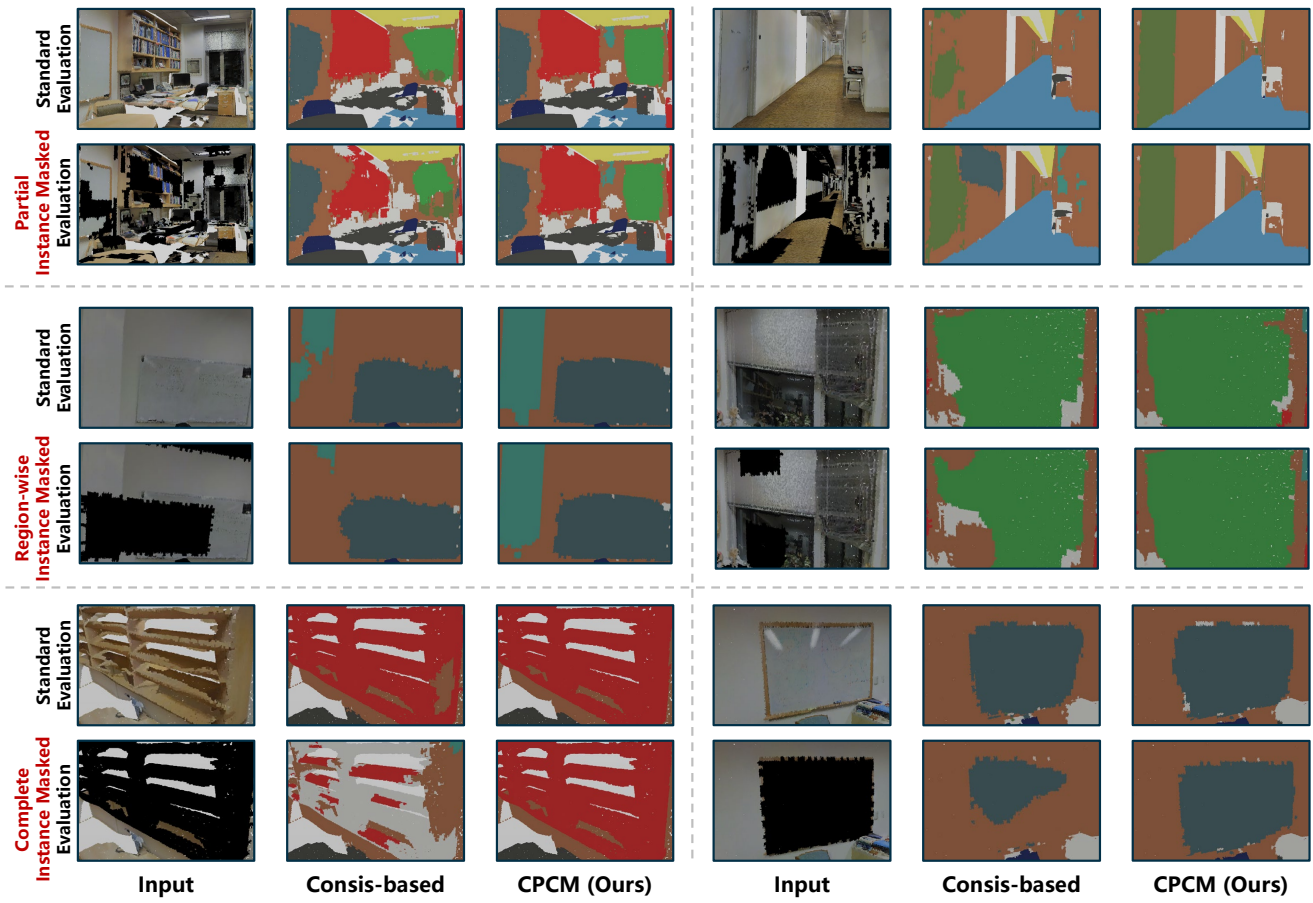

Figure IV: Visual comparison of results from different methods on S3DIS (mask ratio is 40%).

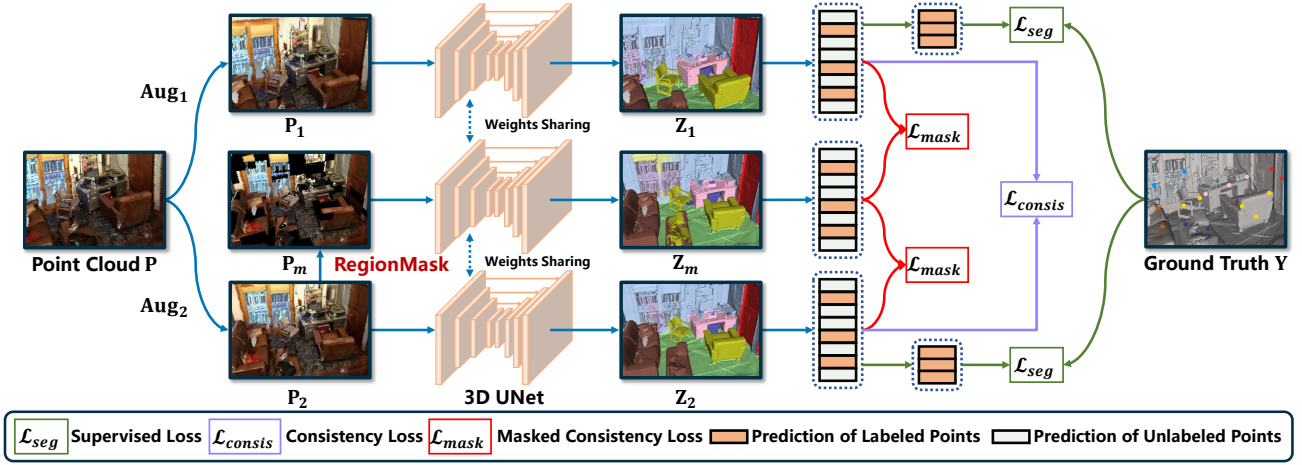

Figure V: Overall scheme of our CPCM method. Given a point cloud $\mathbf{P}$, we first apply two random augmentations and our region-wise masking to obtain the augmented point clouds $\mathbf{P}_1, \mathbf{P}_2$ and the masked point cloud $\mathbf{P}_m$, respectively. Then, the features $\mathbf{Z}_1, \mathbf{Z}_2, \mathbf{Z}_m$ are extracted by a weight-sharing 3D UNet. The supervised cross-entropy loss $\mathcal{L}_{seg}$ is computed over labeled features and a consistency loss $\mathcal{L}_{consis}$ is computed on $\mathbf{Z}_1, \mathbf{Z}_2$. Last, our masked consistency loss $\mathcal{L}_{mask}$ enforces the feature consistency between $\mathbf{Z}_1, \mathbf{Z}_m$ and $\mathbf{Z}_2, \mathbf{Z}_m$ to help the model focus on learning contextual information.

## B. Detailed Formulations of the Loss Functions

For convenience, we show the overall scheme of the proposed CPCM in Figure V. Recall that in Section 3.3 of the main paper, we have introduced the objective of CPCM. Specifically, we propose to learn the masked feature consistency to improve the context comprehension ability of the model. Following the consis-based methods [5, 9], we use the supervised cross-entropy loss on the labeled points and the JS-divergence consistency loss on features from different augmentations. Therefore, the overall objective of the proposed CPCM is defined as

$$\mathcal{L}_{\text{CPCM}} = \mathcal{L}_{seg} + \alpha \mathcal{L}_{consis} + \beta \mathcal{L}_{mask}, \tag{I}$$

where $\mathcal{L}_{seg}, \mathcal{L}_{consis}, \mathcal{L}_{mask}$ indicate the cross-entropy loss, consistency loss and masked consistency loss, respectively. Here, $\alpha$ and $\beta$ are hyper-parameters that control the optimization strength to learn feature consistency across augmentations and contextual information, respectively.

In this section, we provide detailed formulations of the used loss functions. Given differently augmented point clouds $\mathbf{P}_1, \mathbf{P}_2$ from $\mathbf{P}$, the sparse label $\mathbf{Y}$, and the labeled index set $\mathcal{S}$, we first obtain the point-wise classification logits by $\mathbf{Z}_1 = \text{Softmax}(f_\theta(\mathbf{P}_1))$ and $\mathbf{Z}_2 = \text{Softmax}(f_\theta(\mathbf{P}_2))$.

**Cross-entropy loss**. The supervised cross-entropy loss $\mathcal{L}_{seg}$ is computed as follows:

$$\mathcal{L}_{seg} = \frac{1}{|\mathcal{S}|} \sum_{s \in \mathcal{S}} CE(\mathbf{Z}_1[s], \mathbf{Y}[s]) + CE(\mathbf{Z}_2[s], \mathbf{Y}[s]) = -\frac{1}{|\mathcal{S}|} \sum_{s \in \mathcal{S}} \log\big(\mathbf{Z}_1[s]\,[\mathbf{Y}[s]]\big) + \log\big(\mathbf{Z}_2[s]\,[\mathbf{Y}[s]]\big). \tag{II}$$

**Consistency loss**. The consistency loss $\mathcal{L}_{consis}$ is calculated over the unlabeled (or all) points as follows:

$$\mathcal{L}_{consis} = \frac{1}{N} \sum_n JS(\mathbf{Z}_1[n], \mathbf{Z}_2[n]) = -\frac{1}{N} \sum_n \mathbf{Z}_1[n] \log\Big(\frac{\mathbf{Z}'[n]}{\mathbf{Z}_1[n]}\Big) + \mathbf{Z}_2[n] \log\Big(\frac{\mathbf{Z}'[n]}{\mathbf{Z}_2[n]}\Big), \tag{III}$$

$$\mathbf{Z}' = (\mathbf{Z}_1 + \mathbf{Z}_2)/2. \tag{IV}$$

**Masked consistency loss**. Last, given the masked point cloud $\mathbf{P}_m$, we compute its features computed by $\mathbf{Z}_m = \text{Softmax}(f_\theta(\mathbf{P}_m))$ and derive our masked consistency loss $\mathcal{L}_{mask}$ as follows:

$$\mathcal{L}_{mask} = \frac{1}{N} \sum_n JS(\mathbf{Z}'_1[n], \mathbf{Z}_m[n]) + JS(\mathbf{Z}'_2[n], \mathbf{Z}_m[n]), \tag{V}$$

where we stop the gradient flow for the "ground truth" unmasked features $\mathbf{Z}_1, \mathbf{Z}_2$ by the Detach operation in PyTorch, i.e., $\mathbf{Z}'_1 = \text{Detach}(\mathbf{Z}_1), \mathbf{Z}'_2 = \text{Detach}(\mathbf{Z}_2)$.

## C. Technical details on the data augmentation

In this section, we detail the data augmentation used on the point cloud data. We apply the same data augmentation method to get $P_1$ and $P_2$. Following previous methods [4, 7], we use data augmentations including: `RandomDropOut`, `RandomHorizontalFilp`, `ColorAutoContrast`, `ColorTranslation` and `ColorJitter`. The data augmentation is called two times to create differently augmented point clouds to learn feature consistency. Since the augmentation method is the same for $P_1$ and $P_2$, using any one of them to get $P_m$ is feasible.

## D. Effect of hyper-parameters $\alpha$ and $\beta$

In this section, we investigate the effect of hyper-parameters $\alpha$ and $\beta$ that control the optimization strength on consistency loss and mask consistency loss, respectively. We use S3DIS for fast evaluation considering that the number of the training sample is relatively small on S3DIS (204 on S3DIS *vs.* 1201 on ScanNet V2). Moreover, the optimal hyper-parameters for different annotation ratios may vary. Thus, we conduct experiments on S3DIS with annotation ratios 0.01% and 0.1%. We present the experiment results in Table I and our analysis is as follows.

**Effect of hyper-parameter $\alpha$.** The role of $\alpha$ is to control the learning from unlabeled data. As shown in Table I, the optimal value of $\alpha$ is 5 and 1 for the extreme-limited and the limited annotation settings, respectively. Moreover, we observe that the performance does not improve when $\alpha > 1$ under the 0.1% setting. The results indicate that the consistency loss is useful for learning representation as the annotation ratio decreases.

**Effect of hyper-parameter $\beta$.** The hyper-parameter $\beta$ is to control the learning of the masked features prediction task based on unmasked surroundings, which helps the model harness the contextual information in a scene. To better explore the effect of the masked consistency loss, we simply set $\alpha = 0$, which does not apply the consistency loss to learn the weakly-supervised segmentation model. As shown in Table I, the optimal value is 10 and 5 for the extreme-limited and limited annotation setting, which is larger than the optimal value of $\alpha$ *i.e.,* $(\beta, \alpha) = (10, 5)$ when the 0.01% setting and $(\beta, \alpha) = (5, 1)$ for the 0.1% setting. The larger optimal value of $\beta$ indicates that learning contextual information is more effective than learning feature consistency across augmentations. Last, without the consistency loss, the optimal performance of CPCM beat the consis-based method considerably, showing the advantage of considering point contextual relation over point-wise consistency across augmentations only.

Based on the above results, for both S3DIS and ScanNet V2, we simply set $(\alpha, \beta)$ to $(5, 10)$ and $(1, 5)$ for annotation ratio $< 0.1\%$ and $\geq 0.1\%$, respectively. We admit there may be more optimal hyper-parameters by tuning under the specific dataset and annotation settings as well as tuning $(\alpha, \beta)$ simultaneously. For the sake of simplicity, we decide to apply the above coarse settings throughout our experiments.

|  | \multicolumn{4}{c|}{S3DIS 0.01%} | \multicolumn{4}{c}{S3DIS 0.1%} |
| --- | --- | --- | --- | --- | --- | --- | --- | --- |
| $\alpha$ | 1 | 2 | 5 | 10 | 1 | 2 | 5 | 10 |
| CPCM ($\beta = 0$) | 48.6 | 50.7 | **52.9** | 51.8 | **65.0** | 64.9 | 64.7 | 64.3 |
| $\beta$ | 1 | 2 | 5 | 10 | 1 | 2 | 5 | 10 |
| CPCM ($\alpha = 0$) | 51.8 | 53.5 | 56.9 | **59.2** | 65.2 | 65.6 | **66.3** | 64.0 |

Table I: Ablation studies on hyper-parameters $\alpha$ for the Consis-based method and $\beta$ for the proposed CPCM.

## E. Ablation studies on the masked consistency loss $\mathcal{L}_{mask}$

In this section, we provide ablation studies on masked consistency loss. To compute $\mathcal{L}_{mask}$ for $\mathbf{Z}_1$ and $\mathbf{Z}_m$, we align them before the loss calculation. For more details, refer to Section I. As mentioned in Section C, both $\mathbf{P}_1$ and $\mathbf{P}_2$ are the "unmasked" version of $\mathbf{P}_m$. Thus, minimizing the distribution gap between both $\mathbf{Z}_1, \mathbf{Z}_m$ and $\mathbf{Z}_2, \mathbf{Z}_m$ is helpful to learn contextual information. We conduct experiments on the S3DIS dataset with 0.01% annotation. As shown in Table II, using both $JS(\mathbf{Z}_1, \mathbf{Z}_m)$ and $JS(\mathbf{Z}_2, \mathbf{Z}_m)$ achieves the best result.

| $\mathcal{L}_{mask}$ | | mIoU (%) |
|---|---|---|
| $JS(\mathbf{Z}_1, \mathbf{Z}_m)$ | $JS(\mathbf{Z}_2, \mathbf{Z}_m)$ | |
| ✗ | ✗ | 47.7 |
| ✓ | ✗ | 56.5 (+8.8) |
| ✗ | ✓ | 57.2 (+9.5) |
| ✓ | ✓ | **59.3** (+11.6) |

Table II: Ablation studies of $\mathcal{L}_{mask}$ on S3DIS.

## F. Further experiments on the outdoor dataset

To further demonstrate the performance of our CPCM, we provide the quantitative results on the outdoor dataset, SemanticKITTI [2]. Since we followed previous works [4, 7, 8] to use MinkowskiEngine to implement our CPCM, we conduct experiments on the front view part of the SemanticKITTI that provides both XYZ and RGB features for convenience. As shown in Table III, our CPCM consistently provides improvement over the MinkNet and the consis-based baseline. Moreover, thanks to the strong contextual modeling ability, CPCM surpasses the baselines with more annotation, *e.g.*, CPCM (**44.0**, **0.1%**) > consis-based (43.7, 1%) > MinkNet (37.0, 1%). These results demonstrate that our CPCM is able to perform well not only in indoor but also in outdoor scenarios.

| Method | Setting | | |
|---|---|---|---|
| | 1% | 0.1% | 0.01% |
| MinkNet | 37.0 | 30.8 | 23.7 |
| Consis-based baseline | 43.7 (+6.7) | 38.8 (+8.0) | 30.0 (+6.3) |
| CPCM (Ours) | **47.8** (+10.8) | **44.0** (+13.2) | **34.7** (+11.0) |

Table III: Results of mIoU (%) on SemanticKITTI. For reference, the mIoU for fully-supervised MinkNet is 56.4%.

## G. Further experiments on the transformer

To investigate the effectiveness of our CPCM on transformer architecture, we apply CPCM on PTv2 [6], a transformer-based architecture trained in a fully-supervised manner. Since the transformer is generally more data-hungry, we conduct experiments on less weakly-supervised settings (10% or fully supervised) and compare to PTv2. To be specific, we substitute the backbone of CPCM, *i.e.,* MinkNet to PTv2 and the results are shown in Table IV. We observe that our CPCM improves the performance of PTv2 by **5.7%** and **1.6%** with 10% and 100% annotations, respectively. These results verify the effectiveness of CPCM in transformer architecture.

| Settings | PTv2 | PTv2† | PTv2 + CPCM (Ours) |
|---|---|---|---|
| Fully | 71.6 | 69.1 | **70.7** (+1.6) |
| 10% | - | 54.6 | **60.3** (+5.7) |

Table IV: Comparisons with PTv2 on S3DIS, where † denotes the results of our implementation.

## H. Comparison with Unsupervised Pre-training Methods

Unsupervised point cloud pre-training methods can learn useful representations from mass unlabeled data. Thus, existing unsupervised pre-training methods [4, 7] use the pretrained model as initialization and finetune the model on the downstream weakly-supervised point cloud segmentation task. In this section, we investigate the potential of the proposed CPCM by challenging the strong and universal unsupervised pre-training methods: Point Contrast (PC) [7] and Contrastive Scene Context (CSC) [4]. They both leverage point-wise contrastive learning to pre-train the segmentation network. We admit that there are many works on unsupervised point cloud pre-training topics. Since PC and CSC have been evaluated under the weakly-supervised setting, we choose them as our baselines. The results are shown in Table V. The proposed CPCM outperforms PC and CSC under various annotation settings, often by a large margin. Specifically, our CPCM outperforms CSC by 8.87% under the extreme-limited annotation setting 20pts. Moreover, our CPCM train by 50 pts and 100 pts are able to surpass the CSC trained by 100 pts and 200 pts, respectively. Note that our CPCM does not require the pre-training phases and is more suitable for downstream scenarios without large pre-training datasets and computation power.

| Method | 20 pts | 50 pts | 100 pts | 200 pts |
|---|---|---|---|---|
| PC [7] | N/A | N/A | N/A | 67.80 |
| CSC [4] | 53.60 | 60.70 | 65.70 | 68.20 |
| CSC* [4] | 53.80 | 62.90 | 66.90 | 69.00 |
| CPCM (Ours) | **62.67** (+8.87) | **67.89** (+4.99) | **69.67** (+2.77) | **70.32** (+1.32) |

Table V: Comparisons with unsupervised pre-training methods on the ScanNet V2 limited annotated points (pts) per-scene benchmark. * indicates using the active scheme to label representative points.

## I. More Implementation Details

In this section, to facilitate further research, we provide two important implementation details that affect the performance considerably: the point alignment operation and tuning the hyper-parameter weight decay.

**Point alignment.** Note that for both the consistency loss and the masked consistency loss, the alignment operation is required to align two scenes' points before the loss calculation. This is because different augmentations such as random point dropout, geometric clipping and point voxelization would drop some points, which leads to points' misalignment for the two stream data flow. We can resolve this issue by two means: **1) Input level alignment**: all points are aligned *before* feeding into the segmentation network, which leads to the more sparse point cloud data and the loss of some valuable annotations. **2) Feature level alignment**: all points are aligned *after* the feature extraction stage, which causes some feature inconsistency but resolves all issues incurred by the input level alignment solution. We implement both solutions to find out which is better and the results are shown in Table VI. We observe that the feature level alignment strategy performs better w.r.t. different annotation settings and datasets. We attribute the success of feature-level alignment to 1) training and testing under the same input distribution and 2) retraining valuable annotations. Therefore, we choose the feature level alignment as our default point alignment strategy throughout experiments.

| Alignment Strategy | ScanNet V2 | | S3DIS | |
|---|---|---|---|---|
| | 0.01% | 0.1% | 0.01% | 0.1% |
| Input Level | 39.9 | 59.1 | 46.9 | 59.6 |
| Feature Level | **44.2** | **61.8** | **52.9** | **64.9** |

Table VI: Effect of the point alignment position input level *vs.* feature level on the consis-based method.

**Weight decay.** Due to the limited annotation nature in weakly-supervised point cloud segmentation, overfitting is an issue we should consider properly. Thus, we carry out a simple but straightforward way to alleviate the overfitting issue: tuning the weight decay to control the regularization intensity. The results are shown in Table VII. As weight decay increases from $1e^{-4}$ to $1e^{-3}$, both MinkNet and consis-based method achieves better results, which indicates that higher weight decay is able to alleviate the overfitting issue. However, a large weight decay, *i.e.*, $1e^{-2}$ causes the underfitting issue, especially on ScanNet V2 that with thousands of scene point cloud data to be fitted. Thus, we set weight decay to $1e^{-3}$ as our default choice.

| Weight Decay | ScanNet V2 0.01% | | S3DIS 0.01% | |
|---|---|---|---|---|
| | MinkNet | Consis-based | MinkNet | Consis-based |
| $1e^{-4}$ | 36.7 | 41.3 | 45.9 | 50.9 |
| $1e^{-3}$ | **37.6** | **44.2** | **47.7** | **52.9** |
| $1e^{-2}$ | 15.5 | 14.0 | 45.1 | 50.5 |

Table VII: Effect of weight decay on the two baselines: MinkNet and consis-based method.

## J. More Qualitative Results

In this section, we demonstrate the advantage of CPCM with more visualization results in Figure VI. We summed up CPCM's advantage as follows:

**Better at distinguishing adjacent objects**. CPCM is able to disguise geometrically close and appearance similar objects, as shown in row 1 of ScanNet V2 (curtain and wall) and row 4 of S3DIS (door and wall).

**Better at covering the whole object**. CPCM does well in covering large objects as shown in rows 2,4 of ScanNet V2 (bed and table) and rows 1,3 of S3IDS (board and ceiling), indicating CPCM's long-range context comprehension ability.

**Better at recognizing the object with complex structures**. CPCM performs reasonably well at recognizing objects with complex geometric structures and appearance as shown in row 3 of ScanNet V2 (curtain) and row 2 of S3DIS (bookcase).

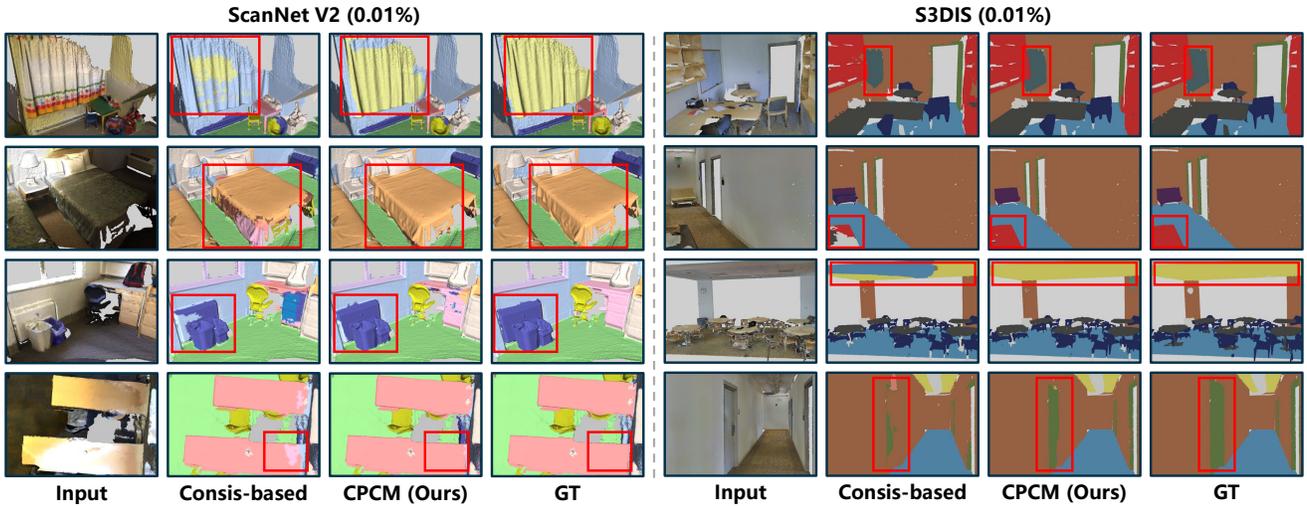

Figure VI: More qualitative comparison between the consis-based method and our CPCM on the ScanNet V2 and S3DIS. We highlight the prediction difference between consis-based method and our CPCM with a red box.

## K. Failure case analysis of CPCM

Our CPCM may fail to effectively distinguish similar classes in the point cloud with a large part of the missing region. Since CPCM heavily relies on the *complete* context to perform accurate segmentation, point clouds with lots of *missing* regions may not provide sufficient context to make correct predictions. For example, in Figure VII, CPCM, unfortunately, hallucinates the door as the window.

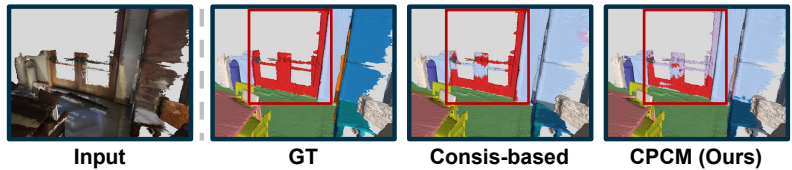

Figure VII: Failure cases of our CPCM on the ScanNet V2.


\end{document}

%% file: def.tex
\usepackage{bm} 

\def\mC{{\mathcal C}}
\def\mD{{\mathcal D}}

\def\mH{{\mathcal H}}

\def\mL{{\mathcal L}}

\def\mS{{\mathcal S}}

\def\mU{{\mathcal U}}

\DeclareMathAlphabet\mathbfcal{OMS}{cmsy}{b}{n}

\def\0{{\bf 0}}
\def\1{{\bf 1}}

\def\bM{{\bf M}}

\def\bP{{\bf P}}

\def\bY{{\bf Y}}
\def\bZ{{\bf{Z}}}

\def\mmR{{\mathbb R}}

\def\bY{{\bf Y}}

\def\bP{{\bf P}}

\def\eg{\emph{e.g.,~}} 
\def\ie{\emph{i.e.,~}} 
 
 \def\vs{\emph{vs.~}}
\def\wrt{{w.r.t.~}}

\def\na{\text{N/A}}
\newcommand{\bestimprove}[2]{\textbf{#1}$_\text{\textcolor{teal}{~(+#2)}}$}
\newcommand{\best}[1]{\textbf{#1}}
\newcommand{\secondimprove}[2]{{#1}$_\text{\textcolor{gray}{~(+#2)}}$}
\newcommand{\noimprove}[2]{{#1}$_\text{\textcolor{red}{~(-#2)}}$}

\usepackage{pifont} 
\newcommand{\with}{\textcolor{green}{\ding{52}}}
\newcommand{\without}{\textcolor{red}{\ding{56}}}